\documentclass{article} %
\usepackage{iclr2027_conference,times}

\usepackage{amsmath,amsfonts,bm}

\def\eqref#1{equation~\ref{#1}}

\def\1{\bm{1}}

\DeclareMathAlphabet{\mathsfit}{\encodingdefault}{\sfdefault}{m}{sl}
\SetMathAlphabet{\mathsfit}{bold}{\encodingdefault}{\sfdefault}{bx}{n}

\definecolor{citecolor}{HTML}{2980b9}
\definecolor{linkcolor}{HTML}{c0392b}
\usepackage[pagebackref=true,breaklinks=true,colorlinks,bookmarks=false,citecolor=citecolor,linkcolor=linkcolor]{hyperref}

\usepackage{url}
\usepackage{booktabs}
\newcommand{\dashedmidrule}{\noalign{\vskip\aboverulesep\hbox to\linewidth{\xleaders\hbox to 4pt{\hss\vrule height\lightrulewidth width 2pt\hss}\hfill}\vskip\belowrulesep}}
\usepackage{etoc}
\usepackage{enumitem}
\usepackage{array}
\usepackage{tabularx}
\newcolumntype{C}{>{\centering\arraybackslash}X}
\usepackage{multirow}
\usepackage{wrapfig}
\usepackage{placeins}
\usepackage{graphicx}
\usepackage{tikz}
\usetikzlibrary{positioning,arrows.meta,calc}
\usepackage{colortbl}
\usepackage{csquotes}
\MakeOuterQuote{"}
\makeatletter\providecommand{\captionof}[1]{\def\@captype{#1}\caption}\makeatother 
\definecolor{oursrow}{HTML}{EDF2F7}   %
\definecolor{grouprow}{HTML}{F3F5EE}  %
\newcommand{\na}{\textcolor{gray}{\textendash}}  %
\newcommand{\yes}{\checkmark}  %
\newcommand{\no}{\textcolor{gray}{$\times$}}  %

\title{Grasping the Future: Joint 3D Hand–Object Forecasting from Internet Videos}

\author{
Rustin Soraki$^{1}$, Matthew Wallingford$^{1}$, Ryan Aby$^{1}$,
Brandon Chee$^{1}$, Jason Ren$^{2}$, \\
\textbf{Ali Farhadi$^{1}$\thanks{Equal supervision.}\kern0.5em,
Homanga Bharadhwaj$^{3}$\footnotemark[2]\kern0.5em,
Roozbeh Mottaghi$^{4}$\footnotemark[2]\kern0.5em} \\
$^{1}$University of Washington \quad
$^{2}$University of North Carolina at Chapel Hill \\
$^{3}$Johns Hopkins University \quad $^{4}$Skild AI
}

\newcommand{\methodname}{\textit{Grasping-the-Future}}
\iclrfinalcopy 
\begin{document}
\etocdepthtag.toc{main}

\setcounter{footnote}{1}
\maketitle

\begin{abstract}
To interact with an object, we need to anticipate both how it will move and how our hands should move with it. Human videos capture these relationships across everyday activities, offering a natural source for learning to predict future interactions. Learning from them requires recovering motion from completed interactions and using it to anticipate what could happen next. We present \methodname{}, a unified generative model of joint 3D hand-object motion, including wrist motion, finger articulation, and object position and orientation. Through masked spatiotemporal completion and flow matching, we train a single model for several prediction tasks: forecasting future interactions, predicting hand motion given object motion or vice versa, generating both trajectories toward a desired final object pose, and completing partially observed interactions. We explicitly model hand-object relations and incorporate contact constraints to help preserve grasp relationships during motion. To obtain 3D supervision from internet videos, we develop a scalable pipeline that combines pretrained perception models with joint geometric optimization to reconstruct completed interactions. We combine these recovered trajectories with annotated interaction datasets as training data for all tasks. We demonstrate strong performance across all tasks, outperforming released forecasters on datasets with 3D annotations. Moreover, we show geometric and physical plausibility of the predicted trajectories by retargeting them to a robotic embodiment. Videos are available at \url{https://grasping-the-future.github.io}.
\end{abstract}

\begin{figure}[h]
\centering
\includegraphics[width=\textwidth]{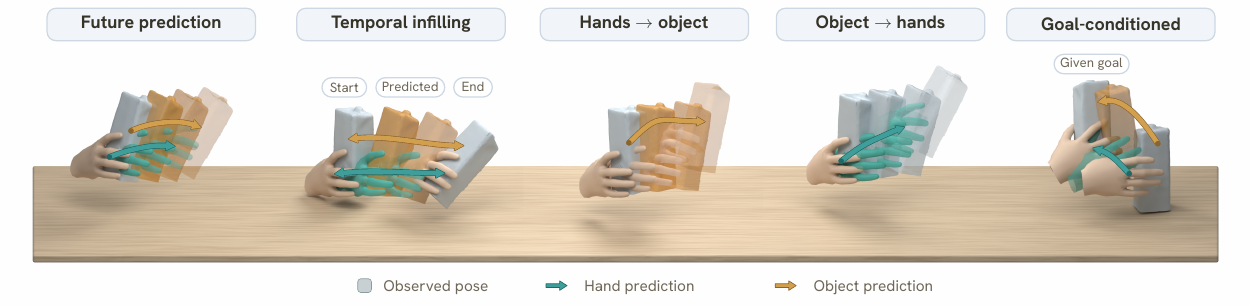}\vspace*{-0.3cm}
\caption{\textbf{Unified model for different types of hand-object prediction.} \methodname{} forecasts future motion, fills temporal gaps, predicts either entity from the other's motion, and generates motion toward a given goal. Neutral poses show observations; teal and amber arrows indicate hand and object prediction, respectively. Fading poses show successive times.}
\label{fig:teaser}\vspace*{-0.3cm}
\end{figure}

\section{Introduction}
\label{sec:introduction}
\begin{flushright}
\small
\textit{``Life can only be understood backwards; but it must be lived forwards.''}\\
--- S{\o}ren Kierkegaard
\end{flushright}

To interact with an object, we need to anticipate how our actions will affect it. Placing a mug on a shelf is a simple everyday action, but predicting how it will unfold requires understanding how the hand and mug move together. The fingers must hold the mug as it is lifted and release it only once it rests on the shelf. The way the mug is grasped affects how it can be carried, and the position of the shelf affects where the hand needs to move. From the eyes of the human manipulating, parts of the hand or the object may be occluded, and several plausible future motions may be possible. Anticipating the interaction therefore requires predicting the motion of both the hand and the object, while preserving firm hand-object contact as the object moves through the 3D world.

Human videos are a natural source for learning these predictions, showing how people handle objects across many everyday situations. Learning from them involves two steps: first, extracting hand and object motion from completed interactions, and then using these examples to learn how an interaction might continue from a partial observation. These steps echo Kierkegaard's distinction: a recorded video allows us to look back and understand what happened, while acting requires anticipating what comes next. Image-based predictions, such as generated videos, can leave relative depth and contact ambiguous. Predicting in 3D makes the spatial relationships between the hands, objects, and scene explicit: where the fingers touch the mug, how it is oriented as it is carried, and where it comes to rest on the shelf. This geometric description also provides an unambiguous reference for later retargeting the predicted motion to a different embodiment like a robot.

Our key insight is that a \textit{single} model can learn these different forms of prediction: predicting an object's trajectory given the hand's motion, predicting the hand's trajectory given the object's motion, generating both trajectories given a desired final object pose, and forecasting or completing an interaction from partial 3D observations. People reason this way intuitively: watching a hand move suggests where an object will go, while imagining where an object should end up suggests how the hand needs to move. We present \textbf{\methodname{}}, a unified generative model of joint 3D hand-object motion. The same model forecasts an interaction from a short observed context, predicts hand trajectories given object motion or object trajectories given hand motion, generates motion toward a desired object pose, and fills gaps between observed poses (Figure~\ref{fig:teaser}). Across these tasks, we must handle observations of different entities and times. We bring these tasks together through masked spatiotemporal completion: we reveal different parts of a hand-object trajectory to the prediction model, and a flow-matching objective learns to generate the missing pieces of the trajectory. The generated trajectories include wrist motion, finger articulation, and object position and orientation. Since plausible hand and object trajectories can still drift apart, we use explicit hand-object relations and contact constraints to keep them moving together during a grasp.

Learning this predictive model requires 3D hand-object supervision that large-scale internet videos do not directly provide. We therefore develop a scalable pipeline that combines pretrained perception models with joint geometric optimization to extract hand-object trajectories from complete monocular videos. The pipeline separates camera motion from object motion, refines the alignment between the hands and object using contact and temporal constraints, and filters unreliable reconstructions. We extract over $62{,}000$ trajectories from internet videos and combine them with existing annotated hand-object datasets in a common 3D representation, accounting for differences in annotation quality during training. On egocentric video, our reconstructions are on par with or slightly better than those of HOPformer~\citep{bansal2026hopformer}, a recent model for egocentric images. From third-person viewpoints, our hand error is about nine times lower and our object pose error about a quarter lower (Appendix~\ref{app:extraction-tables}). The reconstructed trajectories serve as pseudo-ground truth for all tasks.

We evaluate \methodname{} across forecasting and conditional generation tasks on datasets with 3D annotations, including OakInk2~\citep{zhan2024oakink2} and HOT3D~\citep{hot3d}, and on held-out internet videos using our extracted trajectories as pseudo-ground truth. On the annotated datasets, a single forecast from \methodname{} has $17\%$ lower final wrist error and $10\%$ lower final object error than the best released methods. The same model also predicts object motion from hand motion, which none of these methods supports, with $63\%$ lower final error than keeping the object rigidly attached to the hand. We also demonstrate that these predictions can guide robot manipulation: by retargeting the predicted human motion to simulated robot hands with SPIDER~\citep{spider}, we enable the hands to manipulate objects along the forecast trajectories for different types of manipulation tasks.

In summary, our contributions are threefold. We develop a unified generative model of 3D hand-object motion that supports forecasting, cross-entity prediction, goal-conditioned generation, and temporal completion through a common masking formulation. We propose a scalable pipeline for reconstructing temporally aligned hand-object trajectories from internet videos and using them as supervision for learning future interactions. We incorporate explicit hand-object relations and geometric contact constraints into trajectory generation, producing structured 3D predictions that we evaluate through robot retargeting and object manipulation.

\section{Related Work}
\vspace{-0.3cm}
\noindent\textbf{Reconstructing hand-object interactions from video.}
Recent methods recover 3D hand-object geometry and motion from monocular videos by combining visual observations with geometric and learned priors~\citep{hasson2021homan}. Joint priors over hand configurations and object shape support reconstruction and grasp synthesis~\citep{ye2024ghop}, while temporal attention helps recover coherent hand trajectories~\citep{ye2026haptic}. Other methods reconstruct bimanual interactions using Gaussian splatting~\citep{on2025bigs} or use novel-view synthesis priors to recover object geometry that is not visible in short clips~\citep{wang2025magichoi}. Hand priors also support joint hand-object pose estimation from a single egocentric image~\citep{bansal2026hopformer}. Recent work recovers world-space hand and object motion from egocentric videos with known object templates and supports hand-guided interaction synthesis~\citep{ye2026whole}. We build on these advances to extract supervision from completed human interactions. Our focus is on using the recovered trajectories to learn how an interaction unfolds over time from a partial observation.

\noindent\textbf{Interaction forecasting and robot manipulation.}
Prior work predicts hand trajectories and interaction hotspots from egocentric video~\citep{liu2022joint}. Recent methods account for uncertainty in future 3D hand motion~\citep{bao2023usst}, use language queries to guide trajectory prediction~\citep{bao2025handsonvlm}, and combine interaction reasoning with 3D hand motion prediction~\citep{chen2026egoman}. Forecasting object poses~\citep{soraki2026objectforesight} and 3D point trajectories~\citep{bharadhwaj2026motionforesight,molmomotion} further describes how objects move during an interaction. Some methods model the relationship between human and object motion by conditioning object forecasts on hand motion~\citep{thakkar2025polyautoregressive} or combining diffusion with physics-informed correction for full-body human-object forecasting~\citep{xu2023interdiff}. Other work forecasts motion in unconstrained videos through dense 2D point trajectories~\citep{thakkar2026forecasting}. These predictions can also guide robot manipulation. HOPMan translates predicted human interaction plans into robot actions~\citep{bharadhwaj2024hopman}, while Track2Act combines predicted point tracks with geometric transfer and a residual policy~\citep{bharadhwaj2024track2act}. Gen2Act learns policies conditioned on generated human videos~\citep{bharadhwaj2025gen2act}. We learn to predict articulated hand motion together with object pose, capturing how the fingers, wrist, and object move through an interaction. We then retarget these forecasts to simulated robot hands to test their use for manipulation.

\noindent\textbf{Controllable motion synthesis.}
A complementary line of work, often motivated by graphics and animation, generates plausible motion from text, object geometry, reference grasps, or goal poses. These methods typically emphasize generation plausibility without necessarily requiring grounding the predictions in a real scene. Prior work synthesizes dynamic grasps from a grasp reference and a desired object pose using physics-based control~\citep{christen2022dgrasp}. Bimanual manipulation datasets also provide models for completing task primitives~\citep{zhan2024oakink2}. Text-conditioned methods generate hand-object motion through contact-guided diffusion~\citep{cha2024text2hoi} or masked latent flow matching~\citep{chen2026hoflow}, while other work separates global motion from local hand articulation using latent diffusion and a grasp VAE~\citep{li2025latenthoi}. Some methods ground generated motion through initial human-object poses and sparse object waypoints~\citep{li2024chois}. Masks over frames and joints also support sparse or partial motion constraints~\citep{cohan2024flexible}. We ground our predictions in partial observations of a hand-object interaction and the surrounding scene geometry. In \methodname{}, one model forecasts its continuation, predicts either entity from the other's motion, generates both trajectories toward a desired object pose, and completes missing portions of an interaction while modeling hand-object contact.

\vspace{-0.2cm}
\section{Method}
\label{sec:method}
\vspace{-0.2cm}
\begin{figure}[t]
\centering
\includegraphics[width=\textwidth]{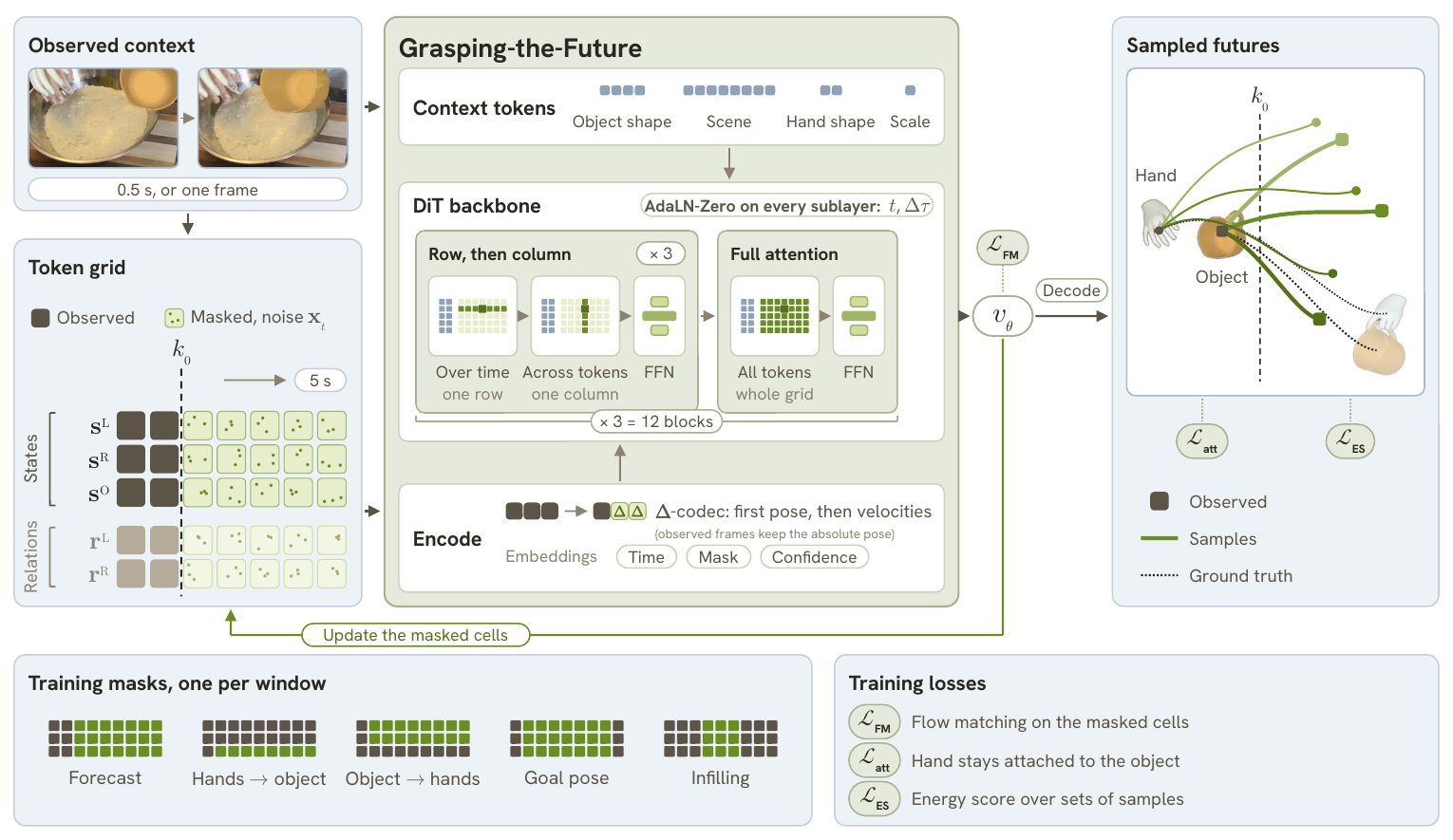}\vspace*{-0.3cm}
\caption{\textbf{Model overview.} An interaction is represented as a grid of tokens, with rows indicating states $\mathbf{s}$ and relations $\mathbf{r}$, and one column per frame. Observed cells are given, and unobserved cells are initialized to noise. A transformer attends to rows, columns and across the grid to predict the flow velocity of the noise cells (with the context tokens (blue) as additional keys/values). The noise is decoded into a sample by ten Euler steps, producing hand and object trajectories. Multiple samples for the same input vs. ground truth (dashed). The bottom row is used for training. An observation mask is sampled per window during training so the same model trains on all tasks (but with a different set of observed cells). Flow matching is used for velocity supervision and auxiliary losses are applied to decoded trajectories.}
\label{fig:model}\vspace*{-0.4cm}
\end{figure}

We develop \methodname{} to predict how hands and objects interact with each other in 3D. We represent an interaction as a trajectory of hand and object poses. We observe a part of this trajectory, and our goal is to generate the rest. By changing which parts we observe, we use the same model to forecast the future, to generate the hand trajectory from a given object trajectory and vice versa, to generate the motion that reaches a given final pose, and to fill a gap in the middle of a trajectory. We first describe this representation and the prediction model and then introduce the losses that help maintain hand-object contact. Finally, we describe how we obtain 3D hand-object reconstructions from internet videos to create a dataset for training the model.

\subsection{Problem Formulation}
\label{sec:formulation}

\noindent\textbf{Representation.}
We segment each recorded interaction into windows of $T$ consecutive frames. A window contains up to two hands and one rigid object, which are the three entities we model. At frame $k$, each hand $e\in\{L,R\}$ has a state $\mathbf{s}^{e}_{k}\in\mathbb{R}^{99}$ made of the wrist position and rotation and the $15$ joint rotations of the MANO hand model~\citep{romero2017embodied}. The object state, $\mathbf{s}^{O}_{k}\in\mathbb{R}^{9}$, is made of its position and rotation. We use the continuous 6-D form~\citep{zhou2019continuity} for the rotations. The states of a window form $\mathbf{X}$. We also give the model a static context $\mathbf{c}$ with the object mesh and its scale, the MANO shape of each hand, the gravity direction when available, and a point cloud of the scene.

\noindent\textbf{Input and output.}
A mask $\mathbf{m}\in\{0,1\}^{3\times T}$ is formed with one row per entity and one column per frame, so each cell represents one entity at one frame. If an entity's state is visible at that frame, a cell is $1$; therefore, its complement $\bar{\mathbf{m}}$ indicates the cells the model needs to generate. The model learns a distribution over the generated states, $\mathbf{X}_{\bar{\mathbf{m}}}\sim p_\theta(\,\cdot\mid\mathbf{X}_{\mathbf{m}},\mathbf{c})$, and samples from this distribution. This formulation unifies the tasks by using randomized masks to train a single model on all of them (Table~\ref{tab:masks}). At evaluation, the masks are:
\begin{itemize}[leftmargin=1.2em,itemsep=1pt,topsep=2pt]
\item \textbf{Forecasting}. The first $0.5$~s of the window are observed and the remaining $4.5$~s are generated for all entities.
\item \textbf{Hands $\to$ object} and \textbf{object $\to$ hands.} One entity is observed throughout the window and the other only at the first frame.
\item \textbf{Goal-conditioned generation.} Only the first frame is observed, together with the final state of the object, the hands, or both, and the model generates the motion toward that final state.
\item \textbf{Temporal infilling.} The middle third of the window is hidden for every entity and generated from the frames before and after it.
\end{itemize}

\noindent\textbf{Window coordinates.}
We call the last frame with an observed cell the present frame. For forecasting, it is the last frame of the observed $0.5$~s. For hands $\to$ object and object $\to$ hands, where one entity is always observed, it is the last frame of the window. We re-express all poses in the camera coordinates of the present frame, which removes camera motion from the window. We also subtract the 3D point halfway between the observed wrists and object at that frame from every position so that each window is centered on the interaction.

\subsection{Generation as Masked Spatiotemporal Completion}
\label{sec:model}

We generate the masked cells with a flow-matching transformer over a grid of tokens (Figure~\ref{fig:model}).

\noindent\textbf{Relation tokens.}
We also include a \emph{relation token} for each hand-object pair, which encourages the model to directly output the hand-object relation rather than depending on the two independently generated poses to somehow result in the hand-object correlation. Each frame then has five tokens, the three states and the two relation tokens. The relation token holds the wrist pose expressed in the object's coordinates and the distance from each of the five fingertips to the object surface. Let $(R^{O}_{k}, \mathbf{p}^{O}_{k})$ denote the object's rotation and position at frame $k$, $(R^{e}_{k}, \mathbf{p}^{e}_{k})$ the wrist's rotation and position, $\mathbf{u}^{e}_{f,k}$ the position of fingertip $f$, and $\mathcal{S}_{k}$ the object surface. In the object's coordinates, the wrist has rotation $Q^{e}_{k}=R_{k}^{O\top}R^{e}_{k}$ and position $\mathbf{q}^{e}_{k}=R_{k}^{O\top}(\mathbf{p}^{e}_{k}-\mathbf{p}^{O}_{k})$. We define the token as
\begin{equation}
\mathbf{r}^{e}_{k}=\Big[\;\mathbf{q}^{e}_{k},\;\;Q^{e}_{k},\;\;\big\{\mathrm{dist}\big(\mathbf{u}^{e}_{f,k},\mathcal{S}_{k}\big)\big\}_{f=1}^{5}\;\Big]\in\mathbb{R}^{14}.
\label{eq:relation}
\end{equation}
If the hand or the object is not observed in a frame, we do not give the model that frame's relation token either, since it would reveal a pose the model has to generate. The model then generates the relation token along with the pose.

\noindent\textbf{$\Delta$-codec and flow matching.}
We represent motion using a temporal-difference encoding, which we call the $\Delta$-codec. The initial state is presented in standardized absolute form and subsequent states as changes per unit time for each continuous interval in which an entity is present. Writing $\mathbf{s}_{k}$ for the state of any entity at time $\tau_{k}$, we encode these changes as
\begin{equation}
\mathbf{z}_{k}=\frac{\mathbf{s}_{k}-\mathbf{s}_{k-1}}{(\tau_k-\tau_{k-1})\,\boldsymbol{\sigma}_{v}}.
\label{eq:codec}
\end{equation}
Here, $\boldsymbol{\sigma}_{v}$ contains per-channel velocity scales estimated from training data. Stationary states have zero difference tokens, and dividing by elapsed time accommodates different frame rates of different windows. A difference token is observed only when both adjacent states are observed. We also add the absolute observed states as additional inputs so the model can attend to observations on both sides of a gap. We reconstruct the absolute states by summing the changes from the interval's start state. During inference, we correct the generated gaps by linear interpolation (in time) of translation corrections and geodesics on SO(3) for rotation corrections of observed entity state past the gap. We preserve all observed states exactly.

We train the masked tokens with flow matching~\citep{lipman2023flow}. With $\mathbf{x}_{1}$ their clean encoded values, $\boldsymbol{\epsilon}$ Gaussian noise and $\mathbf{x}_{t}=(1-t)\boldsymbol{\epsilon}+t\mathbf{x}_{1}$ at a flow time $t\sim\mathcal{U}(0,1)$, the network $v_\theta(\mathbf{x}_{t},t;\mathbf{X}_{\mathbf{m}},\mathbf{c})$ regresses the velocity $\mathbf{x}_{1}-\boldsymbol{\epsilon}$ on the masked tokens, while the observed tokens enter without noise. We call this loss $\mathcal{L}_{\mathrm{FM}}$.

\begin{figure}[t]
\centering
\includegraphics[width=\textwidth]{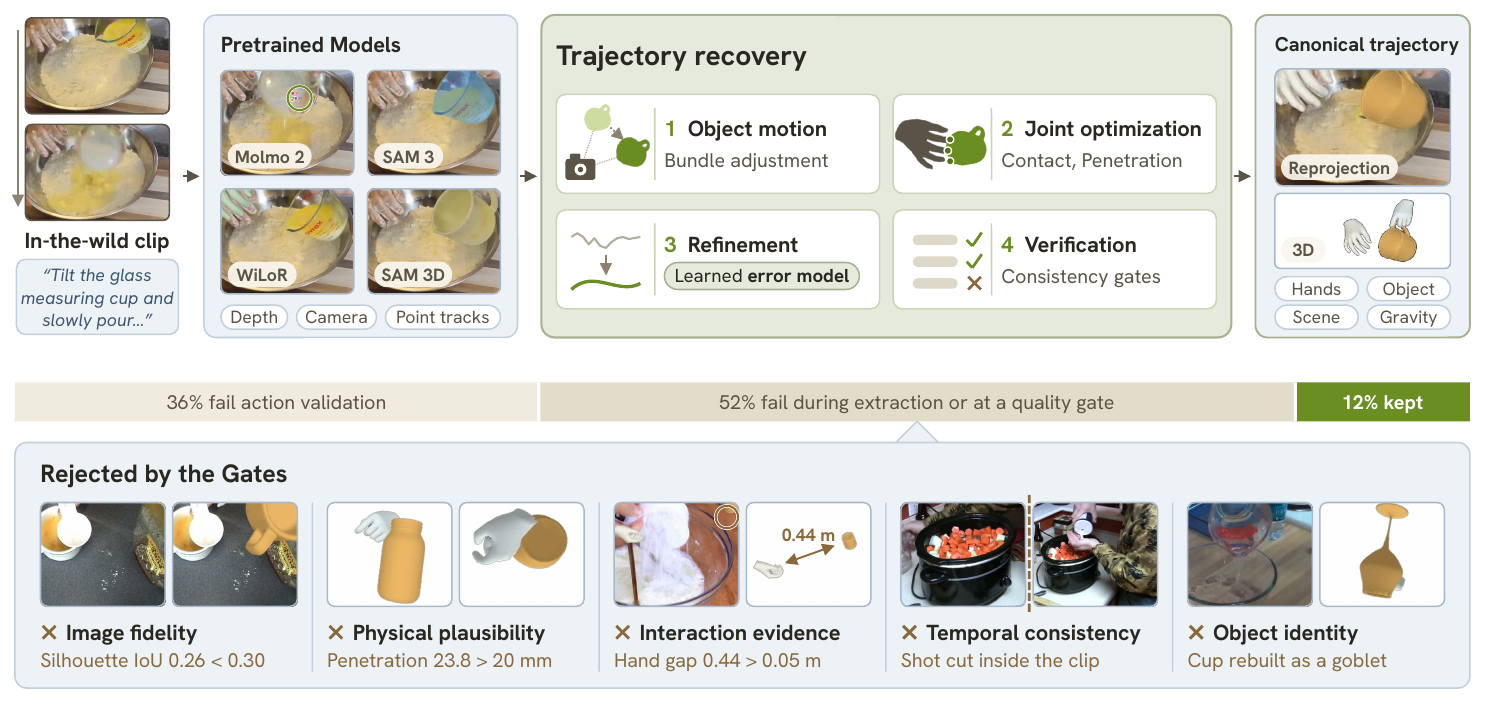}\vspace*{-0.3cm}
\caption{\textbf{Extraction pipeline.} Using four refinement steps, we combine pretrained estimates of hands, objects, depth, and camera motion into a canonical trajectory (green). We only keep clips that pass all consistency checks. The bottom row displays examples rejected by each check, and the bar displays the number of clips we discard across stages.}
\label{fig:pipeline}
\vspace*{-0.4cm}
\end{figure}
\noindent\textbf{Architecture and sampling.}
Each hand, object, or relation token is embedded into the feature space using a layer for its type. Both hands share a layer, as do both relation tokens, so a learned embedding identifies each of the five tokens. We also add a learned embedding for whether the token is observed or masked, a sinusoidal encoding of elapsed time, and a learned projection of its confidence. The confidence distinguishes annotated data from trajectories extracted from video (Section~\ref{sec:extraction}), so the model can rely less on noisy labels. The transformer repeats a group of four blocks. The first three attend to the five tokens over time and to each of the five tokens for each frame. The fourth attends over all tokens of all frames at once. Every attention layer also reads the context $\mathbf{c}$ as extra key-value tokens, with the object mesh encoded by Uni3D~\citep{zhou2024uni3d} and the scene point cloud, carrying DINOv3 features~\citep{dinov3}, by LitePT~\citep{yue2026litept}. At inference, the masked tokens start as Gaussian noise and are denoised by integrating the predicted velocity over $10$ Euler steps.

\subsection{Losses on Decoded Trajectories}
\label{sec:coupling}

Small errors under framewise flow matching can accumulate in decoded trajectories, causing the hand to drift from the object. We therefore supervise hand-object attachment and score complete sampled trajectories.

\noindent\textbf{Attachment loss.}
We estimate and decode a clean trajectory from a single flow-velocity prediction. For each hand $e$, we use the object-frame wrist pose $(Q^{e}_{k},\mathbf{q}^{e}_{k})$ and the object-frame fingertip positions $\mathbf{w}^{e}_{f,k}=R_{k}^{O\top}(\mathbf{u}^{e}_{f,k}-\mathbf{p}^{O}_{k})$. We define reference contact frames $\mathcal{C}_{e}$ as those with at least one fingertip within $5$~mm of the object surface. For each hand, we minimize
\begin{equation}
\mathcal{L}^{e}_{\mathrm{att}}=\sum_{k\in\mathcal{C}_{e}}\Big[\;\|\hat{\mathbf{q}}^{e}_{k}-\mathbf{q}^{e}_{k}\|+\tfrac{1}{5}\textstyle\sum_{f}\|\hat{\mathbf{w}}^{e}_{f,k}-\mathbf{w}^{e}_{f,k}\|\;+\;\lambda_{R}\,d_{\mathrm{c}}\big(\hat{Q}^{e}_{k},Q^{e}_{k}\big)\;+\;\lambda_{v}\,\|\dot{\hat{\mathbf{q}}}^{e}_{k}-\dot{\mathbf{q}}^{e}_{k}\|\;\Big],
\label{eq:attach}
\end{equation}
where hats denote predictions, $d_{\mathrm{c}}$ is the chordal rotation distance, and $\dot{\mathbf{q}}^{e}_{k}$ is the object-relative wrist velocity. We match the reference relative motion, allowing changes such as regrasping. We also supervise fingertip-to-surface distances and penalize acceleration, jerk, and finger-joint violations beyond their reference values (Appendix~\ref{app:model-details}).

\noindent\textbf{Scoring sets of samples.}
To capture multiple plausible futures, we generate $M=3$ trajectories $\hat{\mathbf{X}}_i$ from the same observations on a quarter of optimizer steps. We use six Euler steps per sample and backpropagate through all six. We compare these samples with the annotated or reconstructed trajectory $\mathbf{X}$ using the energy score~\citep{gneiting2007strictly,pacchiardi2024probabilistic,debortoli2025distributional}, which balances agreement with the reference against diversity among samples:
\begin{equation}
\mathcal{L}_{\mathrm{ES}}=\mathrm{ES}\big(\{\phi_{\ell}(\hat{\mathbf{X}}_i)\}_{i=1}^{M},\,\phi_{\ell}(\mathbf{X})\big)\;+\;\lambda_{d}\,\mathrm{ES}\big(\{\phi_{d}(\hat{\mathbf{X}}_i)\}_{i=1}^{M},\,\phi_{d}(\mathbf{X})\big).
\label{eq:es-hybrid}
\end{equation}
The landmark summary $\phi_{\ell}$ records how far each entity, and each wrist relative to the object, has moved after one-third, two-thirds, and all of the generated frames, along with each entity's final rotation change. The dense summary $\phi_d$ records positional changes at every generated frame, with $\lambda_d=0.5$. To maintain focus on final positions and rotations, we score these summaries independently. The expected score is minimized when the predicted and reference summary distributions match~\citep{pic2025aggregation}.

\subsection{Extracting Trajectories from Video}
\label{sec:extraction}

Existing annotated 3D datasets cover a limited range of objects, environments, and interactions, so we also extract hand and object trajectories from monocular internet videos and train on both (Section~\ref{sec:training}). Figure~\ref{fig:pipeline} summarizes the pipeline; we provide implementation details in Appendix~\ref{app:extraction-details} and compare the extracted trajectories with HOPformer on EPIC-Contact, HOT3D, and DexYCB in Appendix~\ref{app:extraction-tables}.

\noindent\textbf{Per-frame estimates.}
We use Molmo~2~\citep{molmo2} to identify manipulation clips from Action100M~\citep{action100m} and locate the manipulated object, then track its mask with SAM~3~\citep{sam3}. We estimate MANO hand poses with WiLoR~\citep{wilor}, object meshes and poses with SAM~3D~\citep{sam3dobjects}, metric depth with MoGe~\citep{moge}, and camera motion with MegaSaM~\citep{megasam}.

\noindent\textbf{Trajectory refinement.}
Each per-frame estimate is in its own frame's camera coordinates, so we use the MegaSaM camera poses to bring them into one world coordinate system. We lift CoTracker3~\citep{cotracker3} object tracks into 3D with depth and fit a rigid object trajectory by robust bundle adjustment~\citep{huber1964robust}. We then jointly refine hand and object poses to match the 2D masks and keep contact, penalizing interpenetration, abrupt motion, and implausible finger poses, with hand scale fixed. When a held object is hidden for a long stretch, its rotation follows the hand's. A transformer trained on corrupted motion-capture trajectories estimates each clip's depth bias, which we correct by shifting the object and the grasping hand along the viewing direction. Finally, we smooth the remaining hand jitter (Appendix~\ref{app:refinement}).

\noindent\textbf{Verification.}
Finally, we check image agreement, contact, interpenetration, temporal consistency, and object size. To make sure we tracked the correct object, we also ask a VLM to point to the object in the hand through the clip. Of the total candidate clips, about $12\%$ yield valid trajectories after final verification. For each verified trajectory, we also compute an interaction-fidelity score. A higher score means the reconstructed hand(s)-object contact fits the video better, the object motion is smoother, and the hands and object align better with their 2D detections. This score goes into the model as part of the confidence input. If a supporting surface is visible, we also estimate the direction of gravity from it.

\begin{figure}[t]
\centering
\includegraphics[width=\textwidth]{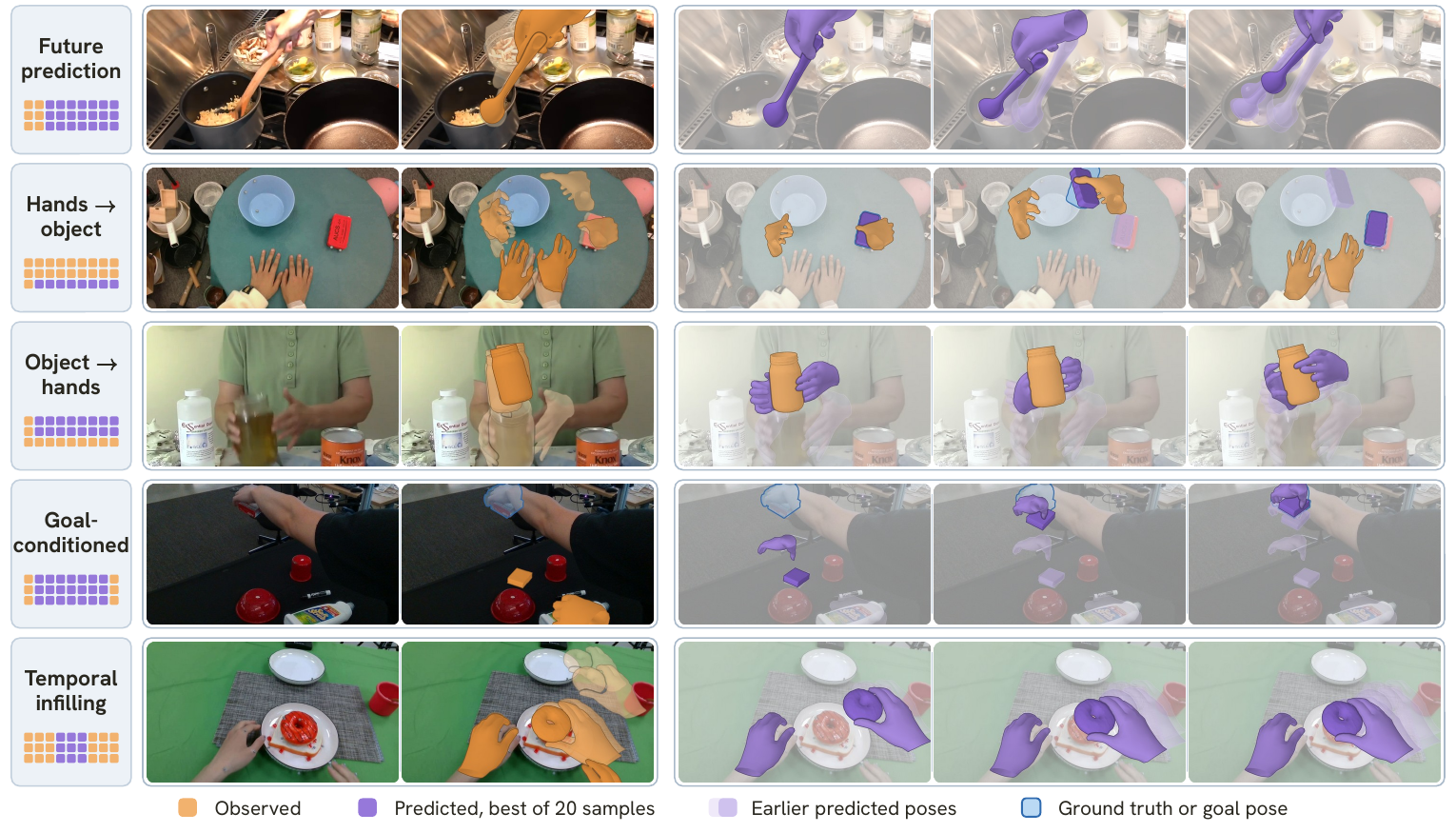}
\vspace*{-0.5cm}
\caption{\textbf{Qualitative predictions across five tasks.} Each row shows one clip, its observation mask, the input frame with and without the observed motion, and three snapshots of the predictions. Mask rows correspond to the two hands and the object. Orange denotes observations and purple predictions; darker orange indicates later observations, while faded purple shows earlier predictions. Blue denotes the reference object pose in the hands-to-object row and the given goal pose.}
\vspace*{-0.3cm}
\label{fig:qual-tasks}
\end{figure}

\section{Experiments}
\label{sec:experiments}
\vspace{-0.3cm}
Our experiments evaluate prediction across seven tasks and its transfer to robotic motion. We compare against non-learned baselines and released forecasting models from prior work, and evaluate the benefit of training on trajectories extracted from internet videos (Section~\ref{sec:exp-tasks}). We then evaluate how well forecast trajectories transfer to robots (Section~\ref{sec:exp-robot}). Additional ablations on the model's losses and inputs are provided in Appendix~\ref{sec:exp-coupling}.

\subsection{Setup}
\label{sec:exp-setup}

\noindent\textbf{Training.}
\label{sec:training}
We train on eight annotated hand-object datasets~\citep{dexycb,hot3d,liu2024taco,zhan2024oakink2,rim2026show3d,hampali2020honnotate,kwon2021h2o,wang2024hocap} and trajectories extracted from internet videos. Together, they provide $74{,}312$ trajectories and $134{,}661$ five-second windows, including $62{,}562$ trajectories and $109{,}346$ windows from internet videos. Each training window receives a task observation pattern sampled uniformly across the five task families, or a random block mask. Training details are provided in Appendix~\ref{app:model-details}.

\noindent\textbf{Benchmark.}
We evaluate every task on the same $331$ five-second windows from diverse scenarios unseen during training, $256$ from the annotated datasets and $75$ from internet videos, with one window per video. For forecasting, the model observes the first $0.5$ seconds and predicts the remaining $4.5$ seconds. For internet videos, the references are our extracted trajectories, which are less accurate than the annotations (Appendix~\ref{app:reference-accuracy}). We will release the code and data publicly.

\noindent\textbf{Metrics.}
We compare each prediction with the recorded trajectory of its window, which we call the reference, and measure position error in millimeters at the wrists and the object center. For a predicted trajectory, the average displacement error (ADE) is this error averaged over its generated frames, while the final displacement error (FDE) is the error at its last frame. When the final pose is observed, the final error is zero by definition, so we report the mid-gap error, calculated at the midpoint of the generated frames. Since a single observation can lead to multiple plausible futures, we generate $20$ samples per input and evaluate the results in three different ways..
\begin{itemize}[leftmargin=1.2em,itemsep=1pt,topsep=2pt]
\item \textbf{Single sample.} We score the first of the $20$ samples on its own, as when the model is asked to make a single prediction.
\item \textbf{Energy score.} The energy score~\citep{gneiting2007strictly,shahroudi2024energy} of the $20$ samples is their average error minus half their average distance from each other. The second term rewards spread, while the first penalizes samples far from the reference. If we view the reference as one draw from all the futures that could follow the observation, the expected score is lowest when the samples come from the same distribution as those futures. A method that makes only one prediction has no spread, so its score equals its error and can be compared directly.
\item \textbf{Best of $\mathbf{20}$.} The error of the sample with the lowest ADE shows how close the nearest of the $20$ samples comes to the reference. This sample is selected using the ground-truth reference, which is unavailable to the model at inference time.
\end{itemize}
\noindent The energy score rewards both accuracy and spread, so we also check the spread on its own. Let $x_1,\dots,x_{20}$ be the sample positions and $x_{21}=y$ the reference position at the frame scored by FDE or mid-gap error, and let $d_i$ be the distance from $x_i$ to the mean of the other $20$ positions. The reference lies inside the samples if $d_{21}$ is not the largest of $d_1,\dots,d_{21}$. If the samples and the reference come from the same distribution, each $d_i$ is equally likely to be the largest, so the reference lies inside the samples in $20/21\approx95\%$ of windows. A rate below $95\%$ means that the samples are too concentrated or centered away from the reference, and a rate above $95\%$ means that they are too dispersed. For all of these metrics, each annotated dataset counts equally, and annotated data and internet videos each count for half. Appendix~\ref{app:benchmark-details} gives full definitions, eligibility rules, and additional interaction metrics.

\noindent\textbf{Baselines.}
Stay-put holds the last observed pose, while constant velocity extrapolates observed motion. Interpolation connects observed endpoint poses. Rigid carry predicts one entity from the other's motion. It keeps their relative pose from the first frame fixed, as if the hand and object were rigidly attached, so a predicted object moves with the nearest wrist and predicted hands move with the object. We also compare with the prior released models for hand and object trajectory forecasting in Table~\ref{tab:exp-external}. Each released model uses its required inputs, and all predictions are evaluated with our scoring code.

\begin{table}[t]
\centering
\vspace{-0.4cm}
\caption{\textbf{Prediction across seven tasks.} Position errors are in millimeters, and hand errors average the left and right wrists. The row “Error at” shows whether a column scores the last frame or, when the final pose is given as input, the middle of the generated frames. “Motion capture only” is our model trained without the trajectories extracted from internet videos. For both models, we report the error of a single sample, the energy score of the $20$ samples, and the error of the best of the $20$ samples (Section~\ref{sec:exp-setup}). Because the energy score subtracts the distance between samples, it is not the error of any one trajectory and can be lower than the best-of-$20$ error. The best of $20$ is chosen using the reference, so it has its own block and is left out when marking the best (bold) and second-best (underlined) results. The last block shows the percentage of windows in which the reference lies within the $20$ samples, ideally $95\%$.}
\label{tab:exp-tasks}
\scriptsize
\setlength{\tabcolsep}{3pt}
\resizebox{\linewidth}{!}{\begin{tabular}{@{}l rr r r rr rr rr rr@{}}
\toprule
& \multicolumn{2}{c}{Future prediction} & Hands$\to$obj. & Obj.$\to$hands & \multicolumn{2}{c}{Temporal infilling} & \multicolumn{2}{c}{Goal, all given} & \multicolumn{2}{c}{Goal, hands given} & \multicolumn{2}{c}{Goal, object given} \\
\cmidrule(lr){2-3} \cmidrule(lr){4-4} \cmidrule(lr){5-5} \cmidrule(lr){6-7} \cmidrule(lr){8-9} \cmidrule(lr){10-11} \cmidrule(lr){12-13}
Entity & object & hands & object & hands & object & hands & object & hands & object & hands & object & hands \\
Error at & end & end & end & end & middle & middle & middle & middle & end & middle & middle & end \\
\midrule
Stay-put & 142.6 & 145.6 & 149.0 & 156.6 & 51.7 & 69.3 & 98.8 & 127.7 & 149.0 & 127.7 & 98.8 & 156.6 \\
Constant velocity & 627.1 & 819.3 & \na & \na & \na & \na & \na & \na & \na & \na & \na & \na \\
Interpolation & \na & \na & \na & \na & 38.2 & 49.0 & 76.0 & 92.3 & \na & 92.3 & 76.0 & \na \\
Rigid carry & \na & \na & 108.6 & 156.4 & \na & \na & \na & \na & \na & \na & \na & \na \\
\midrule
Motion capture only & 159.0 & 185.6 & 71.6 & 132.9 & 39.0 & 50.0 & 91.2 & 101.9 & 146.1 & 110.3 & 96.4 & 166.9 \\
Motion capture only, ES & \underline{115.3} & \underline{136.7} & \underline{52.2} & \underline{97.8} & \underline{25.0} & \underline{34.0} & \underline{61.7} & \underline{67.1} & \underline{99.8} & \underline{69.0} & \underline{62.8} & \underline{117.9} \\
\midrule
\rowcolor{oursrow}Ours & 145.8 & 157.4 & 59.1 & 98.7 & 39.5 & 52.1 & 89.8 & 101.6 & 126.1 & 110.2 & 97.4 & 129.3 \\
\rowcolor{oursrow}Ours, ES & \textbf{89.0} & \textbf{94.4} & \textbf{35.1} & \textbf{59.6} & \textbf{22.2} & \textbf{29.8} & \textbf{51.0} & \textbf{59.3} & \textbf{71.1} & \textbf{62.0} & \textbf{53.5} & \textbf{70.5} \\
\midrule
\rowcolor{grouprow}\multicolumn{13}{@{}l}{Best of the $20$ samples, chosen using the reference}\\
Motion capture only & 121.8 & 153.9 & 50.0 & 103.7 & 26.6 & 35.0 & 62.6 & 74.1 & 107.4 & 78.2 & 68.0 & 137.1 \\
\rowcolor{oursrow}Ours & 102.8 & 113.2 & 36.6 & 75.1 & 23.4 & 32.8 & 56.5 & 70.2 & 91.1 & 73.8 & 61.6 & 91.7 \\
\midrule
\rowcolor{grouprow}\multicolumn{13}{@{}l}{Reference inside the $20$ samples (\%)}\\
Motion capture only & 59 & 48 & 54 & 46 & 72 & 70 & 72 & 65 & 64 & 69 & 75 & 51 \\
\rowcolor{oursrow}Ours & 78 & 76 & 77 & 81 & 86 & 85 & 86 & 79 & 83 & 82 & 87 & 88 \\
\bottomrule
\end{tabular}}
\vspace*{-0.4cm}
\end{table}

\vspace{-0.3cm}
\subsection{Prediction Across Tasks}
\vspace{-0.3cm}
\label{sec:exp-tasks}

Table~\ref{tab:exp-tasks} compares our model with non-learned baselines across all seven tasks. When predicting one entity from the other's motion, a single sample has $46\%$ lower object FDE and $37\%$ lower hand FDE than rigid carry. A single sample is less accurate than stay-put for forecasting ($2$--$8\%$ higher error) and than interpolation at the middle of gaps whose endpoints are both given ($3$--$28\%$). These baselines stay close to the observed poses, which is a safe guess when the future is uncertain. A sample instead commits to one plausible motion, and when this motion differs from the reference, its error can exceed that of not moving at all. The energy score accounts for this by scoring the $20$ samples together, and under it our model scores $30$--$68\%$ lower than the best baseline in every task column. Since the energy score subtracts the distance between samples and a single prediction has none, part of this difference could come from spread alone. To test this, we give stay-put and interpolation the spread of our samples, turning each of their predictions into $20$ samples with Gaussian noise that has the covariance of our samples in that window. With about the same spread, our score is still $9$--$44\%$ lower (Appendix~\ref{app:energy-score}), so our samples also lie closer to the reference. Figure~\ref{fig:qual-tasks} shows examples across five tasks.

We also train the same model without the trajectories extracted from internet videos (“Motion capture only” in Table~\ref{tab:exp-tasks}) to see how much they help. Compared with this model, a single sample of our model has $8\%$ lower forecasting FDE for objects and $15\%$ for hands, and $18$--$26\%$ lower FDE when predicting one entity from the other's motion. When only the final pose of the hands or of the object is given, the FDE of the other entity is $14$--$23\%$ lower. On internet videos, the forecasting gain comes from motion across the image, where extracted references are more accurate than in depth (Appendix~\ref{app:reference-accuracy}). Across the image, the final error falls from $112$ to $73$~mm for objects and from $155$ to $83$~mm for hands, while the error in depth barely changes. The reference also lies inside our samples more often, in $76$--$88\%$ of windows in every column (ideally $95\%$), against $46$--$75\%$ for the motion-capture-only model, whose samples are too close together.

Table~\ref{tab:exp-external} compares our model with released forecasting methods. Under the energy score, our ADE and FDE are $16$--$37\%$ lower than those of the best released method, ForeHand4D for the wrists and ObjectForesight for the object. When each method is scored on one prediction, our wrist ADE and FDE are $13$--$16\%$ lower than those of USST, the most accurate released method in this setting. For the object, MotionForesight has $11\%$ lower ADE and $5\%$ lower FDE than our single sample, but it barely moves the object. Its median predicted displacement at the end of the window is $18$~mm, against $81$~mm for the reference, and its FDE is within $1$~mm of stay-put. The reference lies inside our samples more often than inside those of ObjectForesight, General Flow, and Uni-Hand ($77$--$78\%$ versus $0$--$69\%$). ForeHand4D and MolmoMotion contain it more often ($82\%$ and $95\%$ versus $76\%$ and $71\%$), but their samples are much farther apart and farther from the reference. They are on average $19$ and $48$~cm apart with errors of $22$ and $42$~cm, against $13$ and $11$~cm apart with errors of $16$ and $15$~cm for ours (Table~\ref{tab:es-inside}). A set this wide contains the reference easily but provides poor motion predictions, which the energy score penalizes.

\begin{table}[t]
\centering
\vspace{-0.4cm}
\caption{\textbf{Comparison with different forecasting approaches.} ADE and FDE in millimeters with the listed inputs. Methods with $N$ samples are scored with the energy score of all samples, which equals the error when $N{=}1$. Our model uses estimated depth here. Bold and underline mark the best and second-best results.}
\label{tab:exp-external}
\scriptsize
\setlength{\tabcolsep}{3pt}
\resizebox{\linewidth}{!}{\begin{tabular}{@{}l ccccc c rrr rrr@{}}
\toprule
& \multicolumn{5}{c}{Inputs at inference} & & \multicolumn{3}{c}{ADE} & \multicolumn{3}{c}{FDE} \\
\cmidrule(lr){2-6} \cmidrule(lr){8-10} \cmidrule(lr){11-13}
Method & RGB & Depth & Hands & Object & Text & $N$ & L wrist & R wrist & Object & L wrist & R wrist & Object \\
\midrule
USST \citep{bao2023usst} & \yes & \no & \yes & \no & \no & 1 & 134.6 & 135.2 & \na & 196.3 & 182.3 & \na \\
ForeHand4D \citep{forehand4d} & \yes & \no & \no & \no & \no & 20 & \underline{92.4} & \underline{103.1} & \na & \underline{122.8} & \underline{133.6} & \na \\
EgoH4 \citep{egoh4} & \yes & \no & 2D & \no & \no & 1 & 223.4 & 266.9 & \na & 292.5 & 327.8 & \na \\
Uni-Hand \citep{unihand} & \yes & \yes & \yes & \no & \no & 20 & 360.8 & 344.1 & \na & 359.5 & 348.7 & \na \\
\midrule
ObjectForesight \citep{soraki2026objectforesight} & \yes & \yes & \no & \yes & \no & 20 & \na & \na & \underline{77.0} & \na & \na & \underline{130.4} \\
MotionForesight \citep{bharadhwaj2026motionforesight} & \yes & \yes & \no & mask & \no & 1 & \na & \na & 85.1 & \na & \na & 142.9 \\
General Flow \citep{yuan2024generalflow} & \yes & \yes & \no & 3D pts & \yes & 20 & \na & \na & 121.0 & \na & \na & 198.9 \\
MolmoMotion \citep{molmomotion} & \yes & \no & \no & 3D pts & \yes & 10 & \na & \na & 98.9 & \na & \na & 182.3 \\
\midrule
\rowcolor{oursrow}\methodname{}, one sample & \yes & \yes & \yes & \yes & \no & 1 & 113.1 & 118.0 & 95.9 & 170.3 & 156.1 & 150.9 \\
\rowcolor{oursrow}\methodname{} & \yes & \yes & \yes & \yes & \no & 20 & \textbf{67.5} & \textbf{65.0} & \textbf{55.5} & \textbf{103.4} & \textbf{92.4} & \textbf{91.1} \\
\bottomrule
\end{tabular}
}
\end{table}

\begin{figure}[!t]
\centering
\begin{minipage}[c]{0.66\linewidth}
\includegraphics[width=\linewidth]{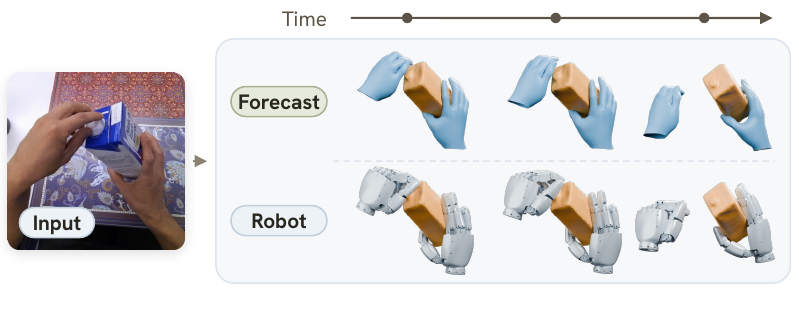}
\end{minipage}\hfill
\begin{minipage}[c]{0.34\linewidth}
\centering
\scriptsize
\setlength{\tabcolsep}{1pt}
\resizebox{\linewidth}{!}{\begin{tabular}{@{}lc@{}}
\toprule
\shortstack[l]{Retargeting method} & \shortstack{Success rate (\%)} \\
\midrule
Kinematic & 14.6 \\
\shortstack[l]{Kinematic + grasp prep} & 43.7 \\
Dynamic (SPIDER) & \textbf{73.8} \\
\bottomrule
\end{tabular}
}
\end{minipage}
\vspace*{-0.5cm}
\caption{\textbf{Testing physical plausibility through retargeting.} We test whether simulated robot hands can manipulate the object along the predicted trajectory from \methodname{}. The left panel shows the observed frame, our $4.5$~s forecast, and simulated replay after SPIDER retargeting~\citep{spider}. The table reports success across $103$ trajectories in different scenes.}\vspace*{-0.3cm}
\label{fig:spider}
\end{figure}

\subsection{Testing Physical Plausibility of Forecasting through Retargeting}
\label{sec:exp-robot}

We assess the physical plausibility of our forecasts by retargeting them to simulated XHands. We test whether the robot hands can manipulate the object and reproduce its predicted motion under the simulator's dynamics. 

Kinematic retargeting matches the robot's wrist and fingertips to the predicted hand poses. Dynamic retargeting with SPIDER~\citep{spider} further optimizes robot controls in simulation to follow the predicted object motion. We also test kinematic retargeting with SPIDER's grasp preparation but without trajectory optimization. Using only the observed frames, grasp preparation adjusts the robot hand's starting pose and grip in simulation until the hand holds the object steadily. While the forecast hand holds the object, it then moves the robot wrist together with the predicted object and keeps the grip unchanged. All three methods use the same forecast for each observation. 

We evaluate $103$ examples whose reference hand-object trajectories retarget successfully with SPIDER, and we run SPIDER with three optimizer seeds per example. We consider a retarget to be successful if the robot hand moves the object through the whole prediction interval and the object's position and rotation errors relative to the forecast, averaged over the predicted frames, are below $100$~mm and $0.5$~rad.

We see that kinematic retargeting succeeds in only $14.6\%$ of examples, showing that these manipulations are difficult to carry out under physics by matching wrist and fingertip poses alone (Figure~\ref{fig:spider}). Grasp preparation raises success to $43.7\%$, so many of these failures come from a grasp that does not hold the object. When the robot controls are also optimized with SPIDER, $73.8\%$ of our forecasts are reproduced in simulation. Once adapted to the robot hand, most of our forecast hand-object motions can therefore be carried out through physical interaction, which indicates that they are physically plausible.

\vspace{-0.3cm}
\section{Conclusion}
\vspace{-0.3cm}
We presented \methodname{}, a unified model for forecasting and completing joint 3D hand-object motion. We combined masked spatiotemporal completion with explicit hand-object relations and contact supervision to predict coordinated motion under different observations and goals. We reconstructed over $62{,}000$ trajectories from internet videos and showed that this supervision improved forecasting beyond training on annotated datasets alone. Under the energy score, our forecasts outperformed all released hand and object forecasting methods by $16$--$37\%$. We also demonstrated that our forecasts could guide object manipulation after retargeting to simulated robot hands, connecting prediction from human videos to robot execution.





\clearpage
\bibliography{iclr2027_conference}
\bibliographystyle{iclr2027_conference}

\newpage
\appendix

\etocdepthtag.toc{supplementary}
\begingroup
\hypersetup{linkcolor=black}
\etocsettagdepth{main}{none}
\etocsettagdepth{supplementary}{subsection}
\etocsetnexttocdepth{subsection}
\etocstandardlines
\etocsettocstyle{%
  {\LARGE\bfseries Supplementary Material\par}
  \vspace{0.7em}
  {\large\itshape Contents\par}
  \vspace{1.5em}
}{}
\makeatletter
\renewcommand*{\l@section}[2]{%
  \addvspace{0.8em}%
  {\bfseries\@dottedtocline{1}{0em}{2.3em}{#1}{#2}}%
}
\renewcommand*{\l@subsection}[2]{%
  \@dottedtocline{2}{2.3em}{3em}{#1}{#2}%
}
\makeatother
\tableofcontents
\endgroup
\clearpage

\section{Model and Training Details}
\label{app:model-details}

In this section we provide additional details on the model, losses and training from Sections~\ref{sec:model}, ~\ref{sec:coupling}, and ~\ref{sec:training}. Also, Tables~\ref{tab:model-hparams} and~\ref{tab:model-losses} provide information on the settings and loss weights.

\noindent\textbf{Representation.}
We divide each trajectory into five-second windows with a $2.5$-second stride, with the last window ending at the trajectory's endpoint. Frames keep their timestamps and cover 5 seconds for each window, regardless of frame rate. So, a window can hold at most $150$ frames for $30$~fps recordings or about $50$ frames for $10$~fps extracted videos. We pad shorter windows (and windows from trajectories less than 5 seconds) and exclude padding and absent entities from attention and losses.

The coordinates are as described in Section~\ref{sec:formulation}, with two additional details. If the gravity direction is known, we also rotate them so that this direction points downward. When the object is unobserved at the present frame, centering uses the mean wrist position alone, and when no hand is observed there, the object alone.

\noindent\textbf{$\Delta$-codec.}
We compute rotation changes by subtracting consecutive 6-D rotation vectors, including those for finger joints. Relation tokens skip this encoding and describe each frame directly. The absolute observed states of Section~\ref{sec:model} are standardized and paired with a flag, $1$ when the state is observed and $0$ otherwise. Both are zero at unobserved frames. They are mapped to a vector added to the state embedding by a linear layer initialized to zero, so they have no effect at the start of training.

\noindent\textbf{Flow-matching loss.}
The flow-matching loss $\mathcal{L}_{\mathrm{FM}}$ is split into six components, for hand translation, wrist rotation, finger articulation, object translation, object rotation, and hand-object relations. Each component is the squared velocity error averaged over its channels and masked frames. Hand and relation components are computed for each side and averaged over the sides with masked frames. The weights in Table~\ref{tab:model-losses} combine the components into a weighted mean for each window, and the loss averages these means across windows. A component with no masked frames in a window is left out of that window's mean, and a window with no masked frames is left out entirely. This averaging prevents over-weighting windows with many masked frames and components with many channels. For losses on decoded trajectories, we estimate clean tokens as $\hat{\mathbf{x}}_1=\mathbf{x}_t+(1-t)v_\theta$ and decode them into poses.

\noindent\textbf{Architecture.}
We use a $12$-block transformer with $512$-dimensional embeddings and eight attention heads, with the attention pattern and token embeddings of Section~\ref{sec:model}. Each token type has a two-layer MLP that embeds it and a linear layer that predicts its flow velocity. Both hands use the same layers, and so do both relation tokens. AdaLN-Zero~\citep{peebles2023dit} gives every attention and feed-forward layer the flow time and the average time between frames. Another MLP takes the three state tokens and predicts fingertip contact, defined as a fingertip-to-object distance below $5$~mm, and is trained with binary cross-entropy. Other settings are given in Table~\ref{tab:model-hparams}.

Each entity's confidence features describe motion smoothness, stability of fingertip-to-object distances, presence in adjacent frames, and source quality. The relation token receives the confidence features of both its hand and the object. Smoothness and fingertip-distance stability are computed from neighboring frames, so close to a masked frame they can reveal the hidden pose. We therefore set them to $0.5$ within $0.1$ seconds of the entity's masked frames. Fingertip distances depend on both the hand and the object, so for this feature we use the masked frames of both.

\noindent\textbf{Conditioning.}
The scene point cloud is derived from the final frame in which all entities are observed. This is because using partially observed frames might reveal a masked entity's pose. We then remove the hand and object pixels using their masks. We then prune background points farther than $0.5$~m from the hand-object midpoint, widening this to $1$ or $2$~m when too few points remain. We then downsample the point cloud to $4{,}096$ points with farthest-point sampling and express it in the present frame's centered coordinates.

Each scene point has $12$ channels to pack the point's position, surface normal, distances to the hands and object, height above the support plane, normal tilt from vertical, curvature, and log density. It also has $32$ channels of PCA-reduced DINOv3 image features~\citep{dinov3}. We then use LitePT-S~\citep{yue2026litept} (pretrained on ScanNet~\citep{dai2017scannet}) to encode the point cloud. Finally, eight learned queries pool the scene encoder's output into eight scene tokens through two attention blocks. We train only the input layer, the output projection, and the pooling blocks while the rest of the scene encoder remains frozen during training. In $20\%$ of training windows, we zero the image features.

We also uniformly sample $10{,}000$ points from the object mesh's surface and use a frozen Uni3D-g encoder~\citep{zhou2024uni3d} to encode them. A trainable projection layer then maps the encoded vector to four object-shape tokens. We also add one context token for the object's log scale and one for each hand's MANO shape.

\begin{table}[t]
\centering
\caption{\textbf{Observation patterns of each task.} $T_v$ is the number of valid frames. $^{\dagger}$Goal-conditioned training gives goals for all entities, only the hands, or only the object, with probabilities $0.5$, $0.25$, and $0.25$. Entities without a goal are observed only in the first frame.}
\label{tab:masks}
\scriptsize
\setlength{\tabcolsep}{4pt}
\begin{tabular}{@{}>{\raggedright\arraybackslash}p{0.18\linewidth}>{\raggedright\arraybackslash}p{0.40\linewidth}>{\raggedright\arraybackslash}p{0.23\linewidth}>{\centering\arraybackslash}p{0.13\linewidth}@{}}
\toprule
Task & Observed in training & Observed at evaluation & Training probability\\
\midrule
Forecasting & first $K$ frames, $K\sim\mathrm{Unif}\{1,\ldots,T_v-1\}$ & first $0.5$~s & $0.19$\\
Hands $\to$ object & hands throughout, object in the first frame & same & $0.19$\\
Object $\to$ hands & object throughout, hands in the first frame & same & $0.19$\\
Goal-conditioned$^{\dagger}$ & first $1$--$5$ and last $1$--$3$ frames & first and last frames & $0.19$\\
Temporal infilling & all except one gap of $20$--$50\%$ of the window & all except the middle third & $0.19$\\
Random blocks & all except one block per selected entity & training only & $0.05$\\
\bottomrule
\end{tabular}
\end{table}

\noindent\textbf{Training masks.}
Table~\ref{tab:masks} lists the observed frames of each task in training and evaluation and how often each task is drawn during training. In temporal infilling, the gap length and start are drawn uniformly, and the first and last frames are always observed. For random blocks, each entity is selected independently with probability $0.5$ and gets one masked segment of random length, from one frame to all but one valid frame, at a random start. The first frame stays observed, and at least one hand or object pose is masked after it.

\noindent\textbf{Trajectory losses.}
We supervise decoded wrist and object positions with squared error and decoded rotations of the wrists, object, and $15$ finger joints of each hand with $\theta^2/2$, where $\theta$ represents the angle to the reference rotation. These losses also include observed frames decoded by the $\Delta$-codec based on predicted differences, in addition to masked frames.

We also regularize smoothness and finger articulation. To improve smoothness, we penalize increases in the squared acceleration and jerk of the wrists and object over the reference within masked segments, with jerk weighted $1/30$ relative to acceleration. For finger articulation, we penalize joint-limit violations and the MANO pose prior when they exceed the reference's. Table~\ref{tab:model-hparams} lists the joint limits.

The fingertip-distance loss is the absolute error in fingertip-to-object distance, counted only for fingertips within $10$~cm of the object in the reference. For each hand, we evaluate it on the last frame where the hand or the object is predicted and on up to $15$ other such frames chosen at random. The attachment loss of Equation~\ref{eq:attach} applies to periods of at least two consecutive frames in which the hand touches the object in the reference. For each hand, we take up to $48$ frames spread across these periods, longest first. Its velocity term also covers steps between observed and predicted frames within a contact period.

Each energy-score term in Equation~\ref{eq:es-hybrid} is
\begin{equation}
\mathrm{ES}(\{\hat{\mathbf{y}}_i\}_{i=1}^{M},\mathbf{y})
=\frac{1}{M}\sum_i\|\hat{\mathbf{y}}_i-\mathbf{y}\|_2
-\frac{1}{2M(M-1)}\sum_{i\ne j}\|\hat{\mathbf{y}}_i-\hat{\mathbf{y}}_j\|_2,
\label{eq:es}
\end{equation}
where $\hat{\mathbf{y}}_i=\phi(\hat{\mathbf{X}}_i)$ and $\mathbf{y}=\phi(\mathbf{X})$ summarize a sample and the reference. Both summaries measure changes from the reference pose at the present frame, and when the hand or the object is predicted, they also include each wrist's change relative to the object. We scale the dense summary so that its size does not depend on the number of predicted frames. Table~\ref{tab:model-hparams} gives this scaling and how the landmark summary represents the final rotation change. We also penalize predicted objects that extend more than $10$~mm below nearby supporting surfaces, such as tabletops.

\noindent\textbf{Training.}
We train for $40{,}000$ steps with AdamW~\citep{loshchilov2019adamw} and evaluate the final checkpoint (see Table~\ref{tab:model-hparams} for the remaining settings). We sample windows with source and label-quality weights, up-weighting windows with a left hand by up to $10\times$ to balance supervision across hands. We also use a curriculum in which extracted videos make up about $23.5\%$ of sampled windows for the first $20{,}000$ steps, then decrease linearly to $2\%$ by step $32{,}000$ and stay there, to focus the end of training on the more accurate annotated data.

For augmentation, with probability $0.5$, we rotate the hands, object, and scene collectively around the vertical axis by a Gaussian angle with standard deviation $0.35$~rad. Independently, with probability $0.5$, we also resample a $30$~fps window at $30$, $15$, or $10$~fps while maintaining the frame timestamps.

\noindent\textbf{Loss weights.}
Since our losses differ greatly in scale, we use gradient norms to set their weights instead of setting them manually. Before training, we set each loss weight so it contributes a target fraction of the flow-matching gradient by measuring its gradient norm on batches from the annotated datasets. Wrist and object position, the energy score, and the support penalty keep a weight of $1$, and the contact-classifier weight rises linearly from about zero to $0.1$ over the first $5{,}000$ steps. Table~\ref{tab:model-losses} lists the resulting weights.

\begin{table}[h]
\centering
\caption{\textbf{Model and training settings.} Dataset sizes give training trajectories / five-second windows.}
\label{tab:model-hparams}
\scriptsize
\begin{tabular}{@{}p{0.27\linewidth}p{0.7\linewidth}@{}}
\toprule
\multicolumn{2}{@{}l}{\textbf{Architecture}}\\
\dashedmidrule
Transformer & $12$ blocks; width $512$; $8$ heads; feed-forward width $2048$; GELU; dropout $0.1$\\
Parameters & $103.3$M trainable; $12.7$M frozen\\
\midrule
\multicolumn{2}{@{}l}{\textbf{Training data}}\\
\dashedmidrule
Dataset sizes & Action100M $62{,}562$/$109{,}346$; TACO $4{,}089$/$6{,}267$; DexYCB $2{,}280$/$2{,}280$; OakInk2 $2{,}213$/$9{,}938$; SHOW3D $1{,}638$/$3{,}552$; HOT3D $1{,}016$/$1{,}016$; H2O $191$/$930$; HO3D $173$/$867$; HO-Cap $150$/$465$; total $74{,}312$/$134{,}661$\\
Data splits & Grouped by recording\\
Source weighting & Sampling weight per dataset $\propto N_s^{1/4}$ at the start of training, where $N_s$ is its number of training windows\\
Label-quality weights & HOT3D, OakInk2 $3$; DexYCB, TACO, H2O, HO3D, HO-Cap $2$; SHOW3D $1.5$; Action100M $1$\\
\midrule
\multicolumn{2}{@{}l}{\textbf{Optimization}}\\
\dashedmidrule
Optimizer & AdamW, $\beta_1=0.9$, $\beta_2=0.999$, weight decay $0.01$\\
Learning rate & Linear warmup to $10^{-4}$ over $1{,}000$ steps, then cosine decay to $10^{-5}$ at step $40{,}000$\\
Gradient clipping & Global norm $1.0$\\
Batch size & $512$ windows; $32$ per GPU\\
Compute & $16$ H100 GPUs; about two days\\
\midrule
\multicolumn{2}{@{}l}{\textbf{Loss settings}}\\
\dashedmidrule
Finger-joint limits & Twist up to $25^\circ$ at finger bases and $15^\circ$ elsewhere; swing up to $135^\circ$\\
Landmark rotation change & Flattened $0.3(R_{\mathrm{ref}}^\top R_{\mathrm{end}}-I)/\sqrt{2}$, where $R_{\mathrm{ref}}$ is the reference rotation at the present frame and $R_{\mathrm{end}}$ the rotation at the last predicted frame\\
Dense-summary scaling & Each position-change sequence is scaled by $\sqrt{3/n}$, where $n$ is its number of predicted frames\\
\bottomrule
\end{tabular}
\end{table}

\begin{table}[h]
\centering
\caption{\textbf{Training loss weights.} The flow-matching weights combine the six components of $\mathcal{L}_{\mathrm{FM}}$, and the other weights scale the auxiliary losses.}
\label{tab:model-losses}
\scriptsize
\begin{tabular*}{\linewidth}{@{\extracolsep{\fill}}lr@{\hspace{2.5em}}lr@{\hspace{2.5em}}lr@{}}
\toprule
\multicolumn{2}{@{}l}{\textbf{Flow matching}} & \multicolumn{2}{@{}l}{\textbf{Decoded trajectory}} & \multicolumn{2}{@{}l@{}}{\textbf{Contact and samples}}\\
\midrule
Hand translation & $0.884$ & Wrist and object position & $1.0$ & Fingertip distance & $0.335$\\
Wrist rotation & $0.738$ & Wrist and object rotation & $0.368$ & Attachment position & $0.178$\\
Finger articulation & $1.112$ & Finger-joint rotation & $0.422$ & Attachment rotation & $0.041$\\
Object translation & $0.730$ & Smoothness & $3.2\times10^{-4}$ & Attachment velocity & $0.121$\\
Object rotation & $0.306$ & Joint limits & $3.35\times10^{-3}$ & Contact classifier & $0.1$\\
Hand-object relations & $0.185$ & MANO pose prior & $5.0\times10^{-4}$ & Energy score & $1.0$\\
 & & & & Support penalty & $1.0$\\
\bottomrule
\end{tabular*}
\end{table}

\clearpage
\section{Extraction Pipeline Details}
\label{app:extraction-details}

\subsection{Pipeline Implementation}
\label{app:pipeline-implementation}

We describe the exported trajectories, processing steps, and quality checks for the video extraction pipeline introduced in Section~\ref{sec:extraction}.

\noindent\textbf{Exported trajectories.}
Each exported trajectory contains frames sampled at approximately $10$~Hz, with their original video timestamps. It includes MANO parameters~\citep{romero2017embodied} for up to two hands, per-frame validity flags for each hand, and the rotation and translation of a reconstructed object mesh. The export also includes an estimated gravity direction when available.

\noindent\textbf{Clip selection and validation.}
We take candidate clips from Action100M's timestamped action captions~\citep{action100m}. Molmo~2~\citep{molmo2} names the manipulated object from each caption, with no predefined object list, and clips where it names none are discarded. The same model is used again, this time on the video, to determine whether a hand is interacting with any objects. In clips that pass, it checks one frame per second to find when the interaction occurs, and these times decide which frames we use to locate the object. Clips are split at detected shot boundaries, and each shot of at least $4$ seconds is processed independently.

\noindent\textbf{Hand and object estimation.}
WiLoR~\citep{wilor} gives MANO hand estimates in each frame. We discard low-confidence detections and short-lived tracks, and reject the clip if no hand track remains. To find the object, we ask Molmo~2~\citep{molmo2} to point to it in a frame using the name given in the clip's caption. We remove outlier points and require the remaining points to lie near a hand whenever one is tracked in that frame. If we can't find the object with enough points in up to four frames, we discard the candidate clip. SAM~3~\citep{sam3} then tracks the object mask forward and backward through the clip, starting from the accepted points.

SAM~3D~\citep{sam3dobjects} then reconstructs the object mesh from one frame, chosen to have a large object mask. Masks much larger than the adjacent frames or multi-part or oddly shaped are given lower priority. At other sampled frames, SAM~3D reuses this mesh to estimate the object's pose and scale. We run the full reconstruction only when this fails for a frame.

\noindent\textbf{Object motion estimation.}
CoTracker3~\citep{cotracker3} tracks points in the object mask, and depth from MoGe~\citep{moge} lifts them into 3D. RANSAC PnP~\citep{fischler1981random,lepetit2009epnp} gives initial object poses, which we refine before removing tracks with high reprojection error. Bundle adjustment then refines the poses and the points' positions within the object together, with fixed camera poses. The Huber loss ~\citep{huber1964robust} assesses agreement with image tracks, while additional penalties promote smooth rotations and maintain the estimated 3D shape. Rotations come from this fit and object centers from SAM~3D's estimates. If no fit is found or if the median per-frame reprojection error is $>12$ pixels, we fall back on the rotation estimates from SAM 3D.

\noindent\textbf{Joint optimization.}
We then fix the hand shape and size and jointly refine the hand and object poses. Although the object shape is fixed, we refine the object size estimated from MoGe depth with a single scale factor per clip. The objective penalizes interpenetration, fits the reprojected object to its mask, and maintains hand contact with the object. Other terms discourage abrupt motion, unnatural finger poses, and large deviations from the initial estimates. We also penalize changes in the hand's position relative to the object and, during grasps, changes in its relative rotation.

We optimize with Adam and a cosine learning-rate schedule in stages. In the initial frames, we align the wrists and fix those poses. We then refine the wrist and object motions, followed by the finger poses, and finally all of them together. A final stage again refines the wrists and fingers in frames with low tracking confidence and poor hand-object alignment by fixing the object and pulling each finger toward its position relative to the palm in nearby reliable frames. After optimization, we correct object rotation during occlusions, remove a learned depth bias, and smooth the remaining hand jitter (Appendix~\ref{app:refinement}).

\noindent\textbf{Object scale and depth correction.}
An object's reconstruction can change size and depth across frames and still match the image mask, since a larger object farther away casts the same silhouette. We correct this when the 95th-percentile object size is at least $1.35$ times the 5th-percentile size. Each frame's mesh is first rescaled to the clip's median object size. Comparing its projected area with the observed mask area then gives a smooth depth correction, leaving out frames where hands cover at least half the projected object. We reject a correction with too few reliable observations or increased depth variation. Any hand in contact with the object shifts along with it, so the contact is maintained.

\noindent\textbf{Interaction intervals.}
We call a frame an interaction frame when the object is visible and a tracked hand is within $5$~cm of its surface. Consecutive interaction frames form intervals, and we merge intervals separated by at most $2$ seconds. We then extend each interval by up to $0.75$ seconds on both sides to include the frames where the hand approaches the object or releases it. Frames where the estimated object pose jumps abruptly are likely reconstruction errors. We leave them out, and no interval is merged or extended across them. We export each interval as a trajectory if it lasts at least $2.5$ seconds and includes at least eight frames. Trajectories from the same clip do not overlap.

\noindent\textbf{Final pose refinement.}
We adjust each trajectory's object position so that the projected silhouette more closely resembles the observed mask. Each modification is restricted to $0.30$~m of translation and $35^\circ$ of rotation. We consider a hand to be grasping an object if it remains within $4$~cm of it for multiple consecutive frames. To maintain their relative stance throughout a grip, we adjust each frame and move the hand with the object. Next, without altering finger articulation, we smooth the hand's rotation in relation to the object during grasps. Lastly, we relocate each hand by up to $8$~mm to minimize residual hand-object penetration.

\noindent\textbf{Gravity estimation.}
We estimate gravity from a supporting surface, such as a table or the floor, in the MoGe depth maps of several trajectory frames. The surface must appear consistently across these frames and lie below the hands, the object, and most of the scene. We favor large surfaces near the interaction, especially one the object rests on. A wall can meet these conditions too, since the hands and the object are in front of it. But for a wall to be the floor, the camera would usually have to be rolled far to the side or pointed sharply upward. People rarely film their hands like that, so we reject these surfaces. There are two more cases that we skip, because in both of them we cannot tell which way is up. A surface the camera sees at a grazing angle could face up or down. A surface facing the camera squarely could be a wall in front of the camera or a table filmed from straight above. Gravity is the normal of the selected surface, pointing away from the interaction; if no surface qualifies, we leave it undetermined.

\noindent\textbf{Quality filtering.}
We keep only trajectories that agree with the video. The projected object must match the outline and size of its tracked mask, and we reject trajectories in which the object changes shape or size or moves abruptly. A hand must also stay within $5$~cm of the object in at least $25\%$ of the frames. Further checks catch shot boundaries, excessive hand-object penetration, and object masks that follow the hand rather than the object. We also drop trajectories where the hand touches the object in 3D but, in the image, almost never touches its mask. Lastly, we ask Molmo~2 to point to “the object held in the hand” in several frames, and we reject the trajectory if it often returns no point or points far from our tracked mask (i.e., we are likely tracking the wrong object).

\subsection{Refinement Details}
\label{app:refinement}

\noindent\textbf{Object rotation.}
When the object is occluded for at least three consecutive frames, the grasping hand's motion guides its rotation only when contact is stable and the camera rotates slowly. During the occlusion, the object rotates more closely with the hand, with the correction growing gradually as the occlusion begins and shrinking as it ends. We reject corrections that introduce large rotation jumps.

\noindent\textbf{Bias denoiser.}
A separate denoiser learns to recover clean trajectories from simulated extraction errors. It trains on $11{,}568$ windows of $50$ frames from DexYCB and HOT3D, sampled at $10$~Hz, excluding the $129$ segments from the sequences we use to evaluate refinement. The simulated errors are designed to reproduce the observed errors in the extraction pipeline. For instance, hands and objects are positioned too near or too far from the camera, their positions are disturbed over several frames, and the object's orientation changes gradually.

The denoiser receives the hand and object poses at each frame, including finger articulation, and predicts corrections that we add to the standardized input states. It is an eight-layer bidirectional transformer with width $384$, six attention heads, and feedforward width $1024$. We train it for $15{,}000$ steps with batch size $256$, learning rate $2\times10^{-4}$, and $500$ warmup steps. The loss is squared reconstruction error with weights of $10$ for translation, $3$ for rotation, and $1$ for finger articulation, plus a velocity term with weight $0.5$ that encourages the recovered motion to match the clean trajectory.

\noindent\textbf{Applying the correction.}
On extracted trajectories, we use only part of this output: how far each hand and the object should move toward or away from the camera. We average these corrections over the trajectory and apply the same shift at each frame. To maintain their relative positions, the object and any hands in contact combine their shifts and move together. Next, we smooth each wrist's position and rotation over three consecutive sampled frames. We do the same for the finger rotations, which we also keep within anatomical limits.

\noindent\textbf{Evaluation.}
Table~\ref{tab:refine-ablation} evaluates refinement on $30$ DexYCB and $28$ HOT3D sequences. Full refinement follows the procedure described above, which keeps only the denoiser's constant depth shift. We compare the final recipe with the raw, unrefined poses and two other refinement variants: one that applies every denoiser correction and one that takes a five-frame moving average of the raw poses. Full refinement lowers the median wrist error from $106.4$ to $88.6$~mm on DexYCB and from $191.2$ to $158.7$~mm on HOT3D. Although applying every denoiser correction reduces depth bias about as much as full refinement, it increases hand surface error on both datasets and object rotation drift on HOT3D. Moving-average smoothing slightly lowers the error on DexYCB but raises it on HOT3D.

\begin{table}[t]
\centering
\caption{\textbf{Refinement results.} Median errors in millimeters, except object rotation drift in degrees. Hand metrics use the right hand, and surface is the hand vertex error after aligning the wrists. Depth errors are signed and best at zero.}
\label{tab:refine-ablation}
\footnotesize
\setlength{\tabcolsep}{3pt}
\begin{tabular}{@{}lrrrrrrrrr@{}}
\toprule
& \multicolumn{4}{c}{DexYCB} & \multicolumn{5}{c}{HOT3D} \\
\cmidrule(lr){2-5} \cmidrule(lr){6-10}
Variant & wrist & depth & surface & obj.\ depth & wrist & depth & surface & obj.\ depth & drift \\
\midrule
Raw extraction & $106.4$ & $-93.5$ & $26.1$ & $-56.2$ & $191.2$ & $153.3$ & $67.4$ & $69.7$ & $30.1$ \\
All denoiser corrections & $97.6$ & $-4.3$ & $41.9$ & $-12.3$ & $162.5$ & $72.5$ & $93.6$ & $19.6$ & $43.5$ \\
Moving-average smoothing & $99.2$ & $-62.4$ & $29.1$ & $-48.6$ & $205.7$ & $173.2$ & $72.1$ & $75.7$ & $32.2$ \\
Full refinement & $88.6$ & $-4.7$ & $27.3$ & $-23.8$ & $158.7$ & $78.6$ & $70.2$ & $19.4$ & $30.1$ \\
\bottomrule
\end{tabular}
\end{table}

\subsection{Extraction Quality}
\label{app:extraction-tables}

We evaluate the extraction pipeline of Section~\ref{sec:extraction} on the EPIC-Contact test set~\citep{bansal2026hopformer} with the authors' released evaluator and fitted reference poses (Table~\ref{tab:ext-epic}). Unlike HOPformer, our pipeline is not trained on EPIC-Contact. It produces trajectories for $215$ of the $237$ test clips, and we score both methods on these clips. Since our trajectories are sampled at about $10$~Hz, we sample HOPformer's predictions at the same frames for metrics that depend on frame rate. Our hand joint error is lower than HOPformer's ($16.0$ vs.\ $19.7$~mm), and so is our motion deviation ($10.6$ vs.\ $13.8$~mm). Our contact deviation ($34.6$ vs.\ $20.8$~mm) and hand-object relative position error ($71.8$ vs.\ $64.7$~mm) are higher, reflecting the main limitation of the extracted trajectories, namely absolute depth and hand placement relative to the object.

\begin{table}[h]
\centering
\caption{\textbf{EPIC-Contact reconstruction.} Published results cover all $237$ test clips. We score the other rows on the $215$ clips our pipeline completes. Motion deviation and acceleration depend on frame rate, so we compute them for both methods at our frames, sampled at about $10$~Hz. Acceleration is given for the hand and the object. SR is the percentage of frames whose object pose error is below $5\%$ or $10\%$ of the object diameter. HOPformer is trained on EPIC-Contact and, like the other published methods, receives prepared hand crops and each clip's ground-truth object mesh. In the last block, HOPformer instead receives only the video frames, hand crops from the WiLoR detector, and the SAM~3D mesh of the object, aligned to the template of the class it predicts, and our pipeline reconstructs the mesh from video given the object's name. Bold marks the better result in the last block.}
\label{tab:ext-epic}
\small
\setlength{\tabcolsep}{4pt}
\resizebox{\linewidth}{!}{\begin{tabular}{@{}lrrrrrrr@{}}
\toprule
Method & \shortstack{Contact\\deviation\,$\downarrow$} & \shortstack{Relative\\position\,$\downarrow$} & \shortstack{Motion\\deviation\,$\downarrow$} & Acceleration\,$\downarrow$ & \shortstack{Hand joint\\error\,$\downarrow$} & SR@0.05\,$\uparrow$ & SR@0.1\,$\uparrow$ \\
 & (mm) & (mm) & (mm) & (m/s$^2$) & (mm) & (\%) & (\%) \\
\midrule
\multicolumn{8}{@{}l}{\textit{Published results}} \\
\addlinespace[2pt]
ArcticNet-SF \citep{fan2023arctic} & 94.2 & 166.7 & 70.1 & 3.9\,/\,5.3 & 44.8 & 16.5 & 55.9 \\
JointTransformer \citep{abouzeid2023jointtransformer} & 30.1 & 78.6 & 20.0 & 3.1\,/\,9.5 & 22.9 & 17.6 & 56.9 \\
HOPformer \citep{bansal2026hopformer} & 20.7 & 65.8 & 11.4 & 2.5\,/\,4.1 & 19.9 & 29.8 & 69.7 \\
\midrule
\multicolumn{8}{@{}l}{\textit{Same $215$ clips, prepared inputs and ground-truth mesh}} \\
\addlinespace[2pt]
HOPformer & 20.8 & 64.7 & 13.8 & 3.9\,/\,7.9 & 19.7 & 29.6 & 68.9 \\
\midrule
\multicolumn{8}{@{}l}{\textit{Same $215$ clips, video frames only}} \\
\addlinespace[2pt]
HOPformer & \textbf{27.2} & \textbf{68.5} & 14.2 & 4.5\,/\,8.1 & 23.9 & 15.1 & 56.1 \\
Ours & 34.6 & 71.8 & \textbf{10.6} & \textbf{3.7}\,/\,\textbf{3.8} & \textbf{16.0} & \textbf{28.5} & \textbf{74.9} \\
\bottomrule
\end{tabular}}
\end{table}

Table~\ref{tab:ext-rgb-only} repeats the comparison when every input comes from the video, as for an internet clip. HOPformer then receives hand crops from the WiLoR detector and the SAM~3D reconstruction of the object, aligned to the template of the class it predicts. On EPIC-Contact, its hand error rises from $19.7$ to $23.9$~mm and its object error from $20.6$ to $24.0$~mm, both above ours ($16.0$ and $18.3$~mm). Its object head was trained to pose its own templates, and it predicts the right class for only $60\%$ of clips and labels almost every DexYCB object a bottle. HOPformer also depends on the egocentric viewpoint. On egocentric HOT3D, its hand error is close to ours ($52.2$ vs.\ $55.0$~mm), but on third-person DexYCB it rises to $96.6$~mm, against $10.3$~mm for ours. The failure depends on the camera. For the camera that views the table from the side, its hand error is $117.8$~mm, compared with $41.4$~mm for a camera that looks down at about $50^\circ$, while ours stays between $9$ and $16$~mm. Our object error is also lower on both datasets ($25.5$ vs.\ $38.5$~mm on HOT3D and $20.7$ vs.\ $28.0$~mm on DexYCB).

\begin{table}[h]
\centering
\caption{\textbf{HOPformer and our pipeline when every input comes from the video.} HOPformer receives the video frames, hand crops from the WiLoR detector, and the SAM~3D mesh of the object, aligned to the template of the class it predicts. Hand error is the $21$-joint error after aligning the wrist. Object error is the ADD-S error after aligning the object centers, as in the EPIC-Contact evaluator, which averages the distance from each point of the predicted mesh to the nearest point of the reference mesh. Errors are in millimeters, and both methods are scored on the same frames. EPIC-Contact is HOPformer's training domain, HOT3D is also egocentric, and DexYCB is filmed from third-person views.}
\label{tab:ext-rgb-only}
\small
\begin{tabular}{@{}lrrrrrr@{}}
\toprule
& \multicolumn{2}{c}{EPIC-Contact} & \multicolumn{2}{c}{HOT3D} & \multicolumn{2}{c}{DexYCB} \\
\cmidrule(lr){2-3}\cmidrule(lr){4-5}\cmidrule(lr){6-7}
Method & Hand $\downarrow$ & Object $\downarrow$ & Hand $\downarrow$ & Object $\downarrow$ & Hand $\downarrow$ & Object $\downarrow$ \\
\midrule
HOPformer & 23.9 & 24.0 & \textbf{52.2} & 38.5 & 96.6 & 28.0 \\
Ours & \textbf{15.8} & \textbf{18.3} & 55.0 & \textbf{25.5} & \textbf{10.3} & \textbf{20.7} \\
\bottomrule
\end{tabular}
\end{table}

HOPformer is cheaper to run. On one H100, it needs about $39$~ms of GPU time per frame including hand detection, or about $9$ GPU-seconds per candidate clip, whereas our pipeline needs about $2.8$ GPU-minutes per candidate clip, or about $28{,}700$ H100-hours for the $610{,}080$ candidate clips we processed. HOPformer, however, can only pose objects from its nine EPIC-Contact classes, has no check that rejects failed reconstructions, and is accurate only near its training viewpoint, so it cannot replace our pipeline for building a corpus from arbitrary internet videos.

\clearpage
\section{Prediction Benchmark Details}
\label{app:benchmark-details}

\noindent\textbf{Benchmark construction.}
The $256$ annotated windows come from $168$ recordings: $36$ from DexYCB; $35$ each from HO3D, HOT3D, OakInk2, SHOW3D, and TACO; $24$ from HO-Cap; and $21$ from H2O. The $75$ internet windows come from $75$ different Action100M videos. We further refine the extracted poses to better match the video and add a new AI-assisted review using six sampled frames to increase reference accuracy. The selection process never uses model predictions or errors, and no window is hand-selected.

\noindent\textbf{How extraction error affects internet-video scores.}
\label{app:reference-accuracy}
We evaluate predictions on internet videos against our own extracted trajectories, so extraction errors can affect the results. Much of what Appendix~\ref{app:refinement} reports, however, is a constant offset per clip, mostly in depth ($45$--$62\%$ of the squared position error on DexYCB, $77\%$ on HOT3D). An offset like this largely cancels. If a clip sits $10$~cm too far from the camera, the observed frames the model receives sit $10$~cm too far as well, and so does the reference. What remains drifts slowly over the clip, is also mostly in depth, and this part does affect the scores. On clips that pass our quality checks, a true motion of $50$--$150$~mm comes out $10$--$15\%$ too short, and the error in that motion is smaller across the image than in depth, with medians of $28$ versus $49$~mm for objects and $41$ versus $55$~mm for wrists.

\noindent\textbf{Trajectory scoring.}
ADE uses only generated frames in which the hand or object is present, and FDE uses only windows in which the reference moves at least $50$~mm from its last observed position. Best of $20$ selects the sample with the lowest ADE summed over the predicted hands and object, and reports every metric for that sample. For samples $x_1,\dots,x_N$ and a reference $y$, the energy score is $\frac{1}{N}\sum_i\lVert x_i-y\rVert-\frac{1}{2N(N-1)}\sum_{i\neq j}\lVert x_i-x_j\rVert$, an unbiased estimate of $\mathbb{E}\lVert X-y\rVert-\frac{1}{2}\mathbb{E}\lVert X-X'\rVert$~\citep{gneiting2007strictly}. It is computed on the same frames and windows as the matching ADE, FDE, or mid-gap error.

\noindent\textbf{Physical metrics.}
Finger error (Table~\ref{tab:exp-components}) is the angle between predicted and reference rotations, averaged over the $15$ finger joints of each hand. For closed object meshes, we measure how deeply hand vertices penetrate the object and the percentage of vertices inside it. Joint limits follow Table~\ref{tab:model-hparams} and are strict, since $14.8\%$ of joint poses in the annotated references exceed them. A violation therefore counts only when the predicted joint exceeds a limit and the reference joint does not, and we report violations as a percentage of all evaluated joint poses.

\noindent\textbf{Interaction metrics.}
The hand-object metrics of Table~\ref{tab:exp-interaction} use frames where the hand and object are both present and at least one of them is generated. A fingertip touches the object when it is within $10$~mm of its surface. Tip error is the fingertip position error relative to the object on frames where the reference hand touches the object, and contact F1 accepts predicted contact up to two frames earlier or later than the reference. A reference grasp is a run of at least three frames in which at least two reference fingertips touch the object. On each grasp frame, the movement error is the difference between how the predicted and reference fingertips have moved relative to the object since the grasp began, averaged over the five fingertips. The frame is held when at least two predicted fingertips touch the object and this error is at most $10$~mm, slipping when a predicted fingertip touches the object but the frame is not held, and detached when every predicted fingertip is at least $20$~mm away. Other frames are unclassified, and we report the percentage of grasp frames in each class.

Each window's interaction score averages the $20$ samples, and the two hands when both are present. We then average windows within each task, separately for annotated data and internet videos, and unlike for the other metrics, every annotated window counts equally regardless of its dataset. Tasks count equally, and so do annotated data and internet videos. For differences between models, we estimate $95\%$ confidence intervals by resampling video segments $5{,}000$ times, using the same resampled segments for both models.

\noindent\textbf{Learned baselines.}
For Table~\ref{tab:exp-external}, we use the USST ResNet-18, ObjectForesight EPIC-DiT, and Uni-Hand SAM-3D-Body/DINOv2 checkpoints, the variant of each that scores best on this benchmark. ObjectForesight takes one step per video frame, which scores better than one step per $1/55$~s, the frame interval of its training videos. The ForeHand4D checkpoint was trained on DexYCB, HOT3D, and H2O, and the Uni-Hand checkpoints on robot demonstrations. Methods that use depth get monocular depth from Depth Anything 3 for MotionForesight and from MoGe-2 for the others, including our model. General Flow and MolmoMotion get action text for each clip, from dataset labels where available and a generic description otherwise. All predictions start from the reference at the last observed frame. To ensure a fair comparison with hand forecasting methods, we shift their predictions to start at the observed wrist, removing a constant placement offset that would otherwise dominate their error. For methods that predict 3D point trajectories, we fit a rotation and translation per frame without changing object size, and we interpolate all predictions to the reference timestamps.

\clearpage
\section{Model Ablations}
\label{sec:exp-coupling}

We ablate our main model with configurations that exclude specific losses, inputs, or augmentations (Tables~\ref{tab:exp-components} and~\ref{tab:exp-interaction}). Each configuration is trained for the same number of steps and with the same optimizer settings as the main model, and we evaluate its final checkpoint as described in Appendix~\ref{app:benchmark-details}.

\begin{table}[!ht]
\centering
\caption{\textbf{Model ablations.} (a) Trajectory errors in millimeters for one sample ($N{=}1$), the best of $20$ ($N{=}20$), and the energy score of all $20$ (ES). ADE averages the six tasks that generate each entity, and FDE the three where its final pose is not given. Inside is the percentage of windows with the reference inside the samples at the final position (Section~\ref{sec:exp-setup}). (b) Physical metrics of one sample.}
\label{tab:exp-components}
\scriptsize
\setlength{\tabcolsep}{1pt}
\textbf{(a) Trajectory accuracy}\par\smallskip
\begin{tabularx}{\linewidth}{l*{4}{C>{\hsize=1.15\hsize}C>{\hsize=.85\hsize}C}CC}
\toprule
& \multicolumn{3}{c}{Hands ADE\,$\downarrow$} & \multicolumn{3}{c}{Object ADE\,$\downarrow$} & \multicolumn{3}{c}{Hands FDE\,$\downarrow$} & \multicolumn{3}{c}{Object FDE\,$\downarrow$} & \multicolumn{2}{c}{Inside (\%)} \\
\cmidrule(lr){2-4} \cmidrule(lr){5-7} \cmidrule(lr){8-10} \cmidrule(lr){11-13} \cmidrule(l){14-15}
Model & $N{=}1$ & $N{=}20$ & ES & $N{=}1$ & $N{=}20$ & ES & $N{=}1$ & $N{=}20$ & ES & $N{=}1$ & $N{=}20$ & ES & Hands & Object \\
\midrule
\rowcolor{oursrow}Main model & 86.7 & 60.2 & 49.6 & 64.9 & 42.0 & 37.5 & 128.5 & 93.3 & 74.8 & 110.3 & 76.9 & 65.1 & 82 & 79 \\
\rowcolor{grouprow}\multicolumn{15}{l}{Losses and relation tokens}\\
No physical-loss group & 90.2 & 62.2 & 51.3 & 64.9 & 43.0 & 38.5 & 135.4 & 101.3 & 80.1 & 111.9 & 80.5 & 68.8 & 81 & 76 \\
No relation tokens & 91.2 & 62.4 & 51.1 & 67.9 & 43.7 & 39.3 & 132.7 & 98.8 & 76.4 & 113.4 & 78.8 & 67.7 & 81 & 76 \\
\rowcolor{grouprow}\multicolumn{15}{l}{Scene conditioning and augmentation}\\
No scene encoder & 86.7 & 60.7 & 50.5 & 66.5 & 42.7 & 38.9 & 129.6 & 95.4 & 75.9 & 116.1 & 80.1 & 69.3 & 84 & 79 \\
No pose or stride augmentation & 90.2 & 62.2 & 51.2 & 67.4 & 43.3 & 39.0 & 134.5 & 102.7 & 78.7 & 115.9 & 81.2 & 69.0 & 80 & 77 \\
\bottomrule
\end{tabularx}

\medskip
\textbf{(b) Physical quality}\par\smallskip
\begin{tabularx}{\linewidth}{l>{\hsize=.8\hsize}C>{\hsize=1.05\hsize}C>{\hsize=1.075\hsize}C>{\hsize=1.075\hsize}C}
\toprule
Model & Finger error ($^\circ$)\,$\downarrow$ & Penetration depth (mm)\,$\downarrow$ & Penetrating vertices (\%)\,$\downarrow$ & Joint-limit violations (\%)\,$\downarrow$ \\
\midrule
\rowcolor{oursrow}Main model & 12.9 & 3.01 & 4.20 & 9.9 \\
\rowcolor{grouprow}\multicolumn{5}{l}{Losses and relation tokens} \\
No physical-loss group & 12.9 & 2.98 & 3.94 & 10.2 \\
No relation tokens & 13.1 & 3.03 & 3.94 & 10.1 \\
\rowcolor{grouprow}\multicolumn{5}{l}{Scene conditioning and augmentation} \\
No scene encoder & 12.8 & 2.93 & 4.14 & 9.9 \\
No pose or stride augmentation & 13.3 & 3.03 & 3.94 & 10.8 \\
\bottomrule
\end{tabularx}
\end{table}

\begin{table}[!ht]
\centering
\caption{\textbf{Interaction quality of the ablations.} Means over the $20$ samples of each window and over all seven tasks. Tip error is the fingertip error during contact, and held, slipping, and detached are percentages of reference-grasp frames. For every ablation, the $95\%$ confidence interval of its difference from the main model excludes zero for tip error, contact F1, and detached frames.}
\label{tab:exp-interaction}
\scriptsize
\setlength{\tabcolsep}{1pt}
\begin{tabularx}{\linewidth}{l*{5}{C}}
\toprule
Model & Tip error (mm)\,$\downarrow$ & Contact F1 (\%)\,$\uparrow$ & Held (\%)\,$\uparrow$ & Slipping (\%)\,$\downarrow$ & Detached (\%)\,$\downarrow$ \\
\midrule
\rowcolor{oursrow}Main model & 65.7 & 68.2 & 15.5 & 47.5 & 24.9 \\
\rowcolor{grouprow}\multicolumn{6}{l}{Losses and relation tokens} \\
No physical-loss group & 72.2 & 65.3 & 14.3 & 44.5 & 29.1 \\
No relation tokens & 70.4 & 66.3 & 14.2 & 45.5 & 28.1 \\
\rowcolor{grouprow}\multicolumn{6}{l}{Scene conditioning and augmentation} \\
No scene encoder & 67.6 & 67.3 & 15.6 & 46.6 & 26.0 \\
No pose or stride augmentation & 70.0 & 66.5 & 14.4 & 45.8 & 27.8 \\
\bottomrule
\end{tabularx}
\end{table}

\paragraph{Accuracy and spread of the samples.}
By energy score, the main model has the lowest ADE and FDE, yielding $1$--$7\%$ lower values than the other configurations for hands and $3$--$6\%$ lower for objects. At the final position, the main model's samples are on average $2$--$6\%$ closer to the reference for hands and $3$--$5\%$ closer for objects, while the distance between samples is about the same or smaller. Across both data sources, the reference lies inside the samples about equally often for all configurations: $80$--$84\%$ of windows for hands and $76$--$79\%$ for objects, versus an ideal of $95\%$. On annotated data, the samples are too close together, and the reference lies inside the main model's samples more often than for the ablations without physical losses, relation tokens, or augmentation ($69\%$ versus $64$--$66\%$ for hands and $69\%$ versus $60$--$63\%$ for objects). On internet video, where every configuration comes close to the ideal, the main model's samples are closer together and closer to the reference.

\noindent\textbf{Losses and relation tokens.}
Two parts of the model are there for hand-object contact, the physical losses and the relation tokens (Section~\ref{sec:model}). The physical losses are the attachment, fingertip-distance, joint-limit, MANO pose prior, and support penalty terms, which supervise contact and plausibility on decoded trajectories (Section~\ref{sec:coupling}). We train one ablation without them and one without the relation tokens. The latter still has the attachment and fingertip-distance losses, so its contact is supervised. Removing either part weakens the grasp. In the main model, fingertip error during contact is $7$--$9\%$ lower, contact F1 is $68.2\%$ (versus $65.3\%$ and $66.3\%$), and $24.9\%$ of grasp frames are detached (versus $29.1\%$ and $28.1\%$; Table~\ref{tab:exp-interaction}). The closer grasp costs a little. Slipping, i.e., frames where the hand touches the object without following the reference grasp, rises to $47.5\%$ (from $44.5\%$ and $45.5\%$), and the share of penetrating vertices to $4.20\%$ (from $3.94\%$), at about the same depth. The main model's hand ADE and FDE are also $2$--$8\%$ lower in every column of Table~\ref{tab:exp-components}.

\noindent\textbf{Scene conditioning.}
Nearby surfaces and obstacles limit where an object can go, and the scene encoder tells the model about them. We also train without it, i.e., without scene geometry or DINOv3 features (the object shape is still given). The hands are barely affected, within $2\%$ on ADE and FDE, and detach about as often ($26.0\%$ versus $24.9\%$ of grasp frames). The objects are not. The main model's object FDE is $4$--$6\%$ lower in every column of Table~\ref{tab:exp-components}, and $7\%$ lower for a single sample on annotated data, where the references are most accurate.

\noindent\textbf{Augmentation.}
We rotate the hands, object, and scene together around the vertical axis and resample windows at lower frame rates (Appendix~\ref{app:model-details}) so that the model does not depend on the usual orientation and frame rate of each training dataset. Without this augmentation, every ADE and FDE in Table~\ref{tab:exp-components} gets worse. The main model's are $3$--$9\%$ lower, and the gap is largest for best-of-$20$ hand FDE. The ablation's hand also detaches from the object more often ($27.8\%$ versus $24.9\%$ of grasp frames). It has the most joint-limit violations in the table ($10.8\%$ versus $9.9\%$) and the largest finger error ($13.3^\circ$ versus $12.9^\circ$).

\noindent\textbf{Interpreting the interaction scores.}
Each sample is scored against a single reference, frame by frame, so a sample that follows a plausible but different hand path is penalized. The scores therefore depend on how much of the hand's path the model has to generate. Fingertip error during contact is $26$~mm in temporal infilling and $46$~mm when the hand motion is given, but $60$--$93$~mm in the five tasks where the model generates most of the hand's path. Table~\ref{tab:exp-interaction} averages all seven tasks. For scale, stay-put on the same frames has a fingertip error of $92$~mm and a contact F1 of $55\%$. The main model beats it on both in every task, and its overall fingertip error is $28\%$ lower.

\clearpage
\section{Additional Model Results}
\label{app:model-results}

This section adds energy-score checks and task-specialist comparisons to the evaluation in Section~\ref{sec:experiments}, following the protocol in Appendix~\ref{app:benchmark-details}.

\subsection{Energy Score}
\label{app:energy-score}
\begin{figure}[!ht]
\centering
{\captionof{table}{\textbf{Energy score with matched spread.} The best baseline of Table~\ref{tab:exp-tasks}, alone and with Gaussian noise matching our samples' spread in each window, against our energy score. Rigid carry is not tested.}
\label{tab:es-control}
\scriptsize
\setlength{\tabcolsep}{3pt}
\resizebox{\linewidth}{!}{\begin{tabular}{@{}l rr r r rr rr rr rr@{}}
\toprule
& \multicolumn{2}{c}{Future prediction} & Hands$\to$obj. & Obj.$\to$hands & \multicolumn{2}{c}{Temporal infilling} & \multicolumn{2}{c}{Goal, all given} & \multicolumn{2}{c}{Goal, hands given} & \multicolumn{2}{c}{Goal, object given} \\
\cmidrule(lr){2-3} \cmidrule(lr){4-4} \cmidrule(lr){5-5} \cmidrule(lr){6-7} \cmidrule(lr){8-9} \cmidrule(lr){10-11} \cmidrule(lr){12-13}
Entity & object & hands & object & hands & object & hands & object & hands & object & hands & object & hands \\
Error at & end & end & end & end & middle & middle & middle & middle & end & middle & middle & end \\
\midrule
Best baseline & 142.6 & 145.6 & 108.6 & 156.4 & 38.2 & 49.0 & 76.0 & 92.3 & 149.0 & 92.3 & 76.0 & 156.6 \\
Best baseline with our spread & 112.2 & 117.2 & \na & \na & 29.9 & 37.8 & 58.6 & 73.1 & 120.2 & 73.2 & 58.6 & 126.5 \\
\rowcolor{oursrow}Ours, energy score of $20$ samples & 89.0 & 94.4 & 35.1 & 59.6 & 22.2 & 29.8 & 51.0 & 59.3 & 71.1 & 62.0 & 53.5 & 70.5 \\
\bottomrule
\end{tabular}
}\par}
\vspace{12pt}
\includegraphics[width=\linewidth]{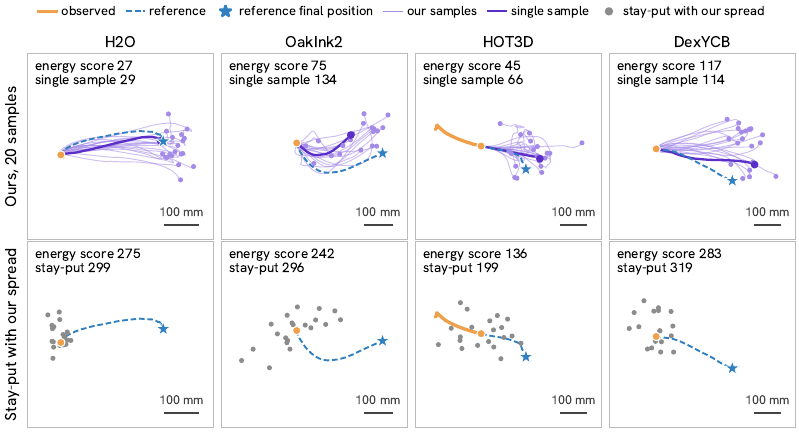}
\caption{\textbf{What the energy score rewards.} Examples for Table~\ref{tab:es-control}. Forecast object paths of our $20$ samples (top) and of stay-put with noise matching their spread (bottom), in the plane of largest motion. The windows are clear cases where our score is lower, which holds in $65\%$ of forecasting windows with at least $50$~mm of object motion. Numbers are in millimeters.}
\label{fig:es-samples}
\end{figure}

Table~\ref{tab:es-control} gives the numbers behind the matched-spread test of Section~\ref{sec:exp-tasks}. With the spread of our samples, the best baseline scores lower than before, but ours is still $9$--$44\%$ lower in every column where the test applies. Paired $95\%$ confidence intervals, from resampling windows within each dataset $5{,}000$ times, exclude zero for every comparison with the best baseline in Table~\ref{tab:exp-tasks}. Figure~\ref{fig:es-samples} shows examples.

Table~\ref{tab:es-inside} applies the inside test of Section~\ref{sec:exp-setup} to the released methods of Table~\ref{tab:exp-external}, at their final predicted positions. USST and MotionForesight make one prediction, so they cannot be tested. The $20$ samples of EgoH4 coincide, which is why Table~\ref{tab:exp-external} lists it with $N{=}1$, and those of Uni-Hand and General Flow almost never contain the reference ($0$--$4\%$ of windows). ObjectForesight contains it in $69\%$ of windows, against $78\%$ for our model. ForeHand4D and MolmoMotion contain it more often than ours, but their samples are much wider and farther from the reference (Section~\ref{sec:exp-tasks}).

\begin{table}[!ht]
\centering
\caption{\textbf{Reference inside the samples.} Forecasting at the final position on the windows of Table~\ref{tab:exp-external}, with both wrists averaged for hand methods. Ours is scored on the same windows and with the same number of samples as the methods in its group. Inside is the percentage of windows with the reference inside the samples, ideally $95\%$ for $N{=}20$ and $91\%$ for $N{=}10$. Apart is the mean distance between two samples and Error the mean distance from a sample to the reference, both in millimeters.}
\label{tab:es-inside}
\scriptsize
\setlength{\tabcolsep}{4pt}
\begin{tabular}{@{}l c rrr@{}}
\toprule
Method & $N$ & Inside & Apart & Error \\
\midrule
\rowcolor{grouprow}\multicolumn{5}{@{}l}{Hand forecasting}\\
ForeHand4D \citep{forehand4d} & 20 & 82\% & 191.8 & 224.1 \\
EgoH4 \citep{egoh4} & 20 & 0\% & 0.2 & 310.2 \\
Uni-Hand \citep{unihand} & 20 & 0\% & 11.2 & 359.8 \\
\rowcolor{oursrow}Ours & 20 & 76\% & 125.2 & 160.5 \\
\midrule
\rowcolor{grouprow}\multicolumn{5}{@{}l}{Object forecasting}\\
ObjectForesight \citep{soraki2026objectforesight} & 20 & 69\% & 107.4 & 184.0 \\
General Flow \citep{yuan2024generalflow} & 20 & 4\% & 32.4 & 215.1 \\
MolmoMotion \citep{molmomotion} & 10 & 95\% & 480.4 & 422.5 \\
\rowcolor{oursrow}Ours & 20 & 78\% & 110.3 & 146.2 \\
\rowcolor{oursrow}Ours & 10 & 71\% & 110.0 & 146.7 \\
\bottomrule
\end{tabular}
\end{table}

\subsection{Task Specialists}
\label{app:task-specialists}
We train five specialists with the shared model's architecture, data, source-sampling schedule, objective, and training budget. Four cover forecasting, hands-to-object prediction, object-to-hands prediction, and temporal infilling, and the fifth covers all three goal patterns. A specialist trains only on its assigned task family and has its own normalization statistics. Table~\ref{tab:exp-tasks-full} compares each one with the shared model on the tasks it trained for.

Under the energy score, the shared model is lower than the specialists in $7$ of the $12$ columns of Table~\ref{tab:exp-tasks-full}, including every forecasting and cross-entity column ($4$--$7\%$). The temporal-infilling specialist is only $2$--$4\%$ lower. The goal specialist gains most for the entity whose final pose is not given, with a $13$--$25\%$ lower energy score. These partial-goal patterns are each about $5\%$ of the shared model's training windows and $25\%$ of the specialist's (Table~\ref{tab:masks}). The reference lies inside the shared model's samples more often in every column, by $2$--$22$ points.

\begin{table}[!ht]
\centering
\caption{\textbf{Comparison with task specialists.} In each column, the specialist is the model trained only for that task, and one goal specialist covers all three goal patterns. Bold marks the better of the two models in each block. Notation follows Table~\ref{tab:exp-tasks}.}
\label{tab:exp-tasks-full}
\scriptsize
\setlength{\tabcolsep}{3pt}
\resizebox{\linewidth}{!}{\begin{tabular}{@{}l rr r r rr rr rr rr@{}}
\toprule
& \multicolumn{2}{c}{Future prediction} & Hands$\to$obj. & Obj.$\to$hands & \multicolumn{2}{c}{Temporal infilling} & \multicolumn{2}{c}{Goal, all given} & \multicolumn{2}{c}{Goal, hands given} & \multicolumn{2}{c}{Goal, object given} \\
\cmidrule(lr){2-3} \cmidrule(lr){4-4} \cmidrule(lr){5-5} \cmidrule(lr){6-7} \cmidrule(lr){8-9} \cmidrule(lr){10-11} \cmidrule(lr){12-13}
Entity & object & hands & object & hands & object & hands & object & hands & object & hands & object & hands \\
Error at & end & end & end & end & middle & middle & middle & middle & end & middle & middle & end \\
\midrule
\rowcolor{grouprow}\multicolumn{13}{@{}l}{Energy score of the $20$ samples}\\
Specialist & 95.2 & 101.2 & 37.5 & 62.0 & \textbf{21.7} & \textbf{28.7} & 53.6 & 60.0 & \textbf{53.0} & \textbf{61.3} & 54.7 & \textbf{61.0} \\
\rowcolor{oursrow}Ours & \textbf{89.0} & \textbf{94.4} & \textbf{35.1} & \textbf{59.6} & 22.2 & 29.8 & \textbf{51.0} & \textbf{59.3} & 71.1 & 62.0 & \textbf{53.5} & 70.5 \\
\midrule
\rowcolor{grouprow}\multicolumn{13}{@{}l}{Reference inside the $20$ samples (\%)}\\
Specialist & 73 & 73 & 58 & 59 & 78 & 78 & 81 & 77 & 80 & 77 & 82 & 83 \\
\rowcolor{oursrow}Ours & \textbf{78} & \textbf{76} & \textbf{77} & \textbf{81} & \textbf{86} & \textbf{85} & \textbf{86} & \textbf{79} & \textbf{83} & \textbf{82} & \textbf{87} & \textbf{88} \\
\bottomrule
\end{tabular}}
\end{table}

\clearpage
\section{Limitations}
\label{app:limitations}

We extract the internet-video trajectories with pretrained models, and their mistakes end up in our training data. MoGe estimates depth from a single image, where a large object far away casts the same silhouette as a small one nearby, so the hands and object can sit at the wrong distance from the camera. Refinement fixes much of this, though not all, and extracted motions come out a bit short. WiLoR loses the fingers when the object covers them and occasionally takes a hand-shaped object for a hand. SAM~3D can flip an object's orientation between frames, and we often fail to pose thin objects such as spoons or cables. On tripod shots, MegaSaM sometimes reports a small camera motion, which then shows up as motion of an object resting on the table. We also reject most candidate clips, many only after all these models and the per-clip optimization have run, so extracting more videos takes a lot of compute.

\end{document}